\def\CleanVersion{1}
\documentclass[journal,final]{IEEEtran}
\usepackage{mathtools, amsmath, amssymb, bbm}
\usepackage{graphicx}
\usepackage{booktabs}
\usepackage{multirow}
\usepackage[switch]{lineno}
\usepackage{arydshln}
\usepackage[dvipsnames]{xcolor}
\usepackage[colorlinks=true, citecolor=BlueGreen, linkcolor=BlueGreen, urlcolor=BlueGreen]{hyperref}
\newcommand{\revision}[1]{#1}
\definecolor{RevisionRed}{RGB}{172,62,62}
\ifdefined\CleanVersion
  \newcommand{\roundtworevision}[1]{\textcolor{black}{#1}}
\else
  \newcommand{\roundtworevision}[1]{\textcolor{RevisionRed}{#1}}
\fi
\newcommand{\tablestyle}{%
  \normalfont\rmfamily\footnotesize
  \renewcommand{\arraystretch}{1.15}%
  \setlength{\tabcolsep}{3pt}%
}

\newcommand{\etal}{\textit{et al}. }
\newcommand{\ie}{\textit{i}.\textit{e}., }

\begin{document}
\title{\roundtworevision{JSolver}: Joint Spectrum Estimation and Multi-Material Decomposition from Single-Energy CT Projections}
\author{Qing~Wu, Hongjiang~Wei,~\IEEEmembership{Member, IEEE}, Jingyi~Yu,~\IEEEmembership{Fellow, IEEE}, S.~Kevin~Zhou,~\IEEEmembership{Fellow, IEEE}, and Yuyao~Zhang,~\IEEEmembership{Member, IEEE}
\thanks{This work was supported by the National Natural Science Foundation of China under Grants 62571328, W2431046, by MoE Key Lab of Intelligent Perception and Human-Machine Collaboration (ShanghaiTech University), and by the Shanghai Frontiers Science Center of Human-centered Artificial Intelligence.}
\thanks{Qing Wu and Jingyi Yu are with School of Information Science and Technology, ShanghaiTech University, Shanghai, China (e-mail: wuqing2025@gmail.com; yujingyi@shanghaitech.edu.cn).}
\thanks{Hongjiang Wei is with School of Biomedical Engineering and National Engineering Research Center of Advanced Magnetic Resonance Technologies for Diagnosis and Therapy (NERC-AMRT), Shanghai Jiao Tong University, Shanghai, China (e-mail: hongjiang.wei@sjtu.edu.cn).}
\thanks{S. Kevin Zhou is with School of Biomedical Engineering \& Suzhou Institute for Advanced Research, University of Science and Technology of China, Suzhou, China (e-mail: skevinzhou@ustc.edu.cn).}
\thanks{Yuyao Zhang (\textit{corresponding author}) is with School of Information Science and Technology, State Key Laboratory of Advanced Medical Materials and Devices, and iHuman Institute, ShanghaiTech University, Shanghai, China (e-mail: zhangyy8@shanghaitech.edu.cn).}}

\maketitle
\begin{abstract}
Multi-material decomposition (MMD) enables quantitative reconstruction of tissue compositions in the human body, supporting a wide range of clinical applications. However, traditional MMD typically requires spectral CT scanners and pre-measured X-ray energy spectra, significantly limiting clinical applicability. To this end, various methods have been developed to perform MMD using conventional (\ie single-energy, SE) CT systems, commonly referred to as SEMMD. Despite promising progress, most SEMMD methods follow a two-step image decomposition pipeline, which first reconstructs monochromatic CT images using algorithms such as FBP, and then performs decomposition on these images. The initial reconstruction step, however, neglects the energy-dependent attenuation of human tissues, introducing severe nonlinear beam hardening artifacts and noise into the subsequent decomposition. This paper proposes \roundtworevision{JSolver}, a fundamentally reformulated one-step SEMMD framework that jointly reconstructs multi-material compositions and estimates the energy spectrum directly from SECT projections. By explicitly incorporating physics-informed spectral priors into the SEMMD process, \roundtworevision{JSolver} accurately simulates a virtual spectral CT system from SE acquisitions, thereby improving the reliability and accuracy of decomposition. Furthermore, we introduce implicit neural representation (INR) as an unsupervised deep learning solver for representing the underlying material maps. The inductive bias of INR toward continuous image patterns constrains the solution space and further enhances estimation quality. Extensive experiments on both simulated and real CT datasets show that \roundtworevision{JSolver} outperforms state-of-the-art SEMMD methods in accuracy and computational efficiency.
\end{abstract}

\section{Introduction}
\par Multi-material decomposition (MMD)~\cite{mendoncca2010multi, long2014multi, mendoncca2013flexible, xue2019accurate} is an important imaging technique that enables the quantitative reconstruction of multiple tissue compositions within the human body. It plays a critical role in a wide range of clinical applications~\cite{rassouli2017detector, leng2019photon}, including virtual non-contrast (VNC) imaging, metal artifact reduction, liver fat quantification, and bone mineral density assessment. However, traditional MMD methods typically rely on spectral (\ie multi-energy) CT scanners and pre-measured X-ray energy spectra, which are often unavailable in most clinical settings~\cite{xue2021multi}, particularly in less-developed regions, thereby significantly limiting their practical use.
\par \revision{To reduce the dependence on multi-energy CT systems, \roundtworevision{single-energy multi-material decomposition (SEMMD)} aims to recover multiple material compositions from conventional SECT data. This problem is highly ill-posed because only one spectral measurement is available. Existing SEMMD methods generally follow a two-step image-domain pipeline. A CT image is first reconstructed from SECT projections, followed by material decomposition. The first step does not explicitly preserve the energy-dependent attenuation information in polychromatic measurements. Beam-hardening artifacts and noise can therefore propagate into the decomposed material maps. These limitations motivate a one-step projection-domain approach that jointly models polychromatic acquisition and material decomposition.}
\par In this paper, we propose \roundtworevision{JSolver}, a fundamentally reformulated one-step SEMMD framework. By explicitly incorporating physics-informed spectral priors into the SEMMD process, \roundtworevision{JSolver} accurately simulates a virtual energy-dependent spectral CT system from SE acquisitions. This enables the simultaneous reconstruction of multi-material compositions and the underlying X-ray energy spectrum directly from SECT projections. This one-step pipeline eliminates the reliance on FBP-like reconstruction algorithms, thereby preserving the intrinsic variations in tissue attenuation across the X-ray energy spectrum and effectively avoiding beam hardening artifacts. Consequently, our method significantly improves SEMMD reconstruction quality. Technically, our \roundtworevision{JSolver} makes three key contributions. First, we formulate a novel one-step optimization framework grounded in rigorous X-ray CT physics, where SEMMD is defined as solving for the volume fractions of multiple materials under the ideal solution assumption, in line with prior MMD works~\cite{mendoncca2010multi,long2014multi,mendoncca2013flexible}. Second, for spectrum estimation, we \revision{introduce a SoftMax transformation into well-established spectrum library-based models~\cite{zhao2014indirect,zhao2016segmentation, zhao2017segmentation, chang2019spectrum}, enabling unconstrained optimization while strictly enforcing non-negative weights that sum to one}. Third, we introduce implicit neural representation (INR) as a powerful unsupervised DL-based solver to reconstruct the material volume fraction maps. The inherent spectral bias of INR toward low-frequency image structures~\cite{rahaman2019spectral} effectively regularizes the ill-posed inverse problem, resulting in high-quality decompositions.
\par We evaluate the performance of the proposed \roundtworevision{JSolver} on both simulated XCAT phantom data and real-world CT data acquired using a research cone-beam CT system and a commercial United Imaging Healthcare (UIH) uCT 768 scanner. Experimental results demonstrate that \roundtworevision{JSolver} consistently outperforms existing state-of-the-art SEMMD methods regarding both reconstruction accuracy and computational efficiency. \revision{\textbf{\textit{To the best of our knowledge, this is the first work to introduce an unsupervised INR-based framework into the domains of spectrum estimation and SEMMD}}}.
\par The main contributions of this work are as follows: 
\begin{itemize}
    \item We propose a novel one-step SEMMD framework that jointly reconstructs material compositions and estimates the X-ray energy spectrum from SECT projections.
    \item \revision{We present the first unsupervised INR-based SEMMD approach that incorporates X-ray physical models, effectively enhancing reconstruction performance.}
    \item \revision{We introduce the SoftMax transformation into spectrum estimation, enabling unconstrained optimization while strictly enforcing the physical constraints on the spectrum weights.}
    \item We conduct extensive experiments demonstrating the superiority of the proposed method over existing SEMMD approaches in both accuracy and efficiency.
\end{itemize}

\section{Related Works}
\par In this section, we provide a brief review of prior studies closely related to our work, including: single-energy multi-material decomposition (Sec.~\ref{sec:semmd}), X-ray CT spectrum estimation (Sec.~\ref{sec:spectraum_estimation}), and implicit neural representation for CT reconstruction (Sec.~\ref{sec:inr_for_ct}).
\subsection{Single-Energy Multi-Material Decomposition}
\label{sec:semmd}
\par \revision{Existing SEMMD methods mainly include optimization-based and supervised DL-based approaches. Optimization-based methods incorporate CT physics and solve the material fractions iteratively. TMA~\cite{xue2020image} assumes a two-material composition and performs voxel-wise optimization, while MSC~\cite{xue2021multi} introduces material sparsity regularization. More recently, Chen et al.~\cite{chen2025multibasis} investigated multi-basis-image reconstruction from conventional CT data. Supervised DL methods learn mappings from SECT images to spectral images or material maps using paired training data. Li et al.~\cite{li2023quality} further introduced quality checking and physical constraints into single-kV material basis estimation. These methods generally operate in the image domain and require either iterative image-domain optimization or paired spectral CT/material labels. In contrast, \roundtworevision{JSolver} reconstructs material maps directly from SECT projections and jointly estimates the unknown spectrum without external training data.}
\subsection{X-ray CT Spectrum Estimation}
\label{sec:spectraum_estimation}
\par X-ray energy spectra describe the distribution of photons emitted by X-ray sources across different energy levels. Accurate spectrum estimation is essential for understanding and modeling nonlinear effects (e.g., beam hardening) in X-ray CT imaging. Existing computational methods for spectrum estimation can be roughly divided into two categories: 1) Physics-based methods~\cite{yang2006robust,chang2016statistical,chang2024parametric}, which use physical models to characterize properties of X-ray generation and interaction, such as bremsstrahlung radiation, photoelectric attenuation, and characteristic emission. As a result, spectrum estimation is formulated as a problem of solving for the underlying physical parameters; 2) Library-based methods~\cite{zhao2014indirect,zhao2016segmentation,zhao2017segmentation,chang2019spectrum}, which leverage a spectrum library consisting of numerous pre-defined standard spectra to represent the unknown energy spectrum. The estimation is then formulated as solving for the optimal combination of weights of these spectra. \revision{For example, Ha~\etal~\cite{ha2019spectrum} estimated CT spectra through Kullback--Leibler divergence constrained optimization.} While both categories have shown promising results, they estimate the spectrum independently from the MMD process and typically require dual-energy or spectral CT acquisitions,  leading to increased acquisition costs and introducing potential cumulative errors in the subsequent decomposition. In contrast, our method integrates spectrum estimation directly into the SEMMD optimization, enabling accurate spectrum recovery from SECT projections while simultaneously achieving one-step multi-material decomposition.
\subsection{\revision{Implicit Neural Representation for CT Reconstruction}}
\label{sec:inr_for_ct}
\par Implicit neural representation (INR) is an unsupervised DL-based framework for solving inverse imaging problems. By incorporating physical forward models into coordinate-based neural networks, INR enables high-quality image reconstruction directly from partial measurements in an unsupervised manner. In recent years, numerous studies~\cite{shen2022nerp, wu2023self, ruckert2022neat, du2024dper, zang2021intratomo, reed2021dynamic, parkimplicit} have explored INR-based approaches for undersampled CT reconstruction, typically adopting the linear projection model as the CT forward operator and demonstrating promising results. More recently, Wu~\etal~\cite{wu2023unsupervised, wu2024solving} extended INR to nonlinear CT imaging by incorporating a polychromatic forward model to account for the X-ray beam hardening effect, thereby effectively reducing metal artifacts. In addition, several preliminary works~\cite{smith2024spectral, shi2024ray} have shown the potential of INR for spectral CT reconstruction. However, the broader applicability of INR to complex nonlinear CT problems remains largely underexplored. In this work, we conduct an in-depth investigation of INR in the challenging nonlinear CT task of joint spectrum estimation and SEMMD, providing new insights into its potential for advancing fundamental CT imaging applications.

\section{Proposed \roundtworevision{JSolver}}
\par In this section, we introduce our \roundtworevision{JSolver} model. First, we formally present a new one-step optimization objective for addressing joint spectrum estimation and SEMMD (Sec.~\ref{sec-Problem Formulation}). Then, we propose two effective representations for the spectrum and material volume fractions (Sec.~\ref{sec:Representations for Spectrum and Volume Fractions}). Finally, we outline the pipeline for optimizing our \roundtworevision{JSolver} (Sec.~\ref{sec:optimization}) and provide its implementation details (Sec.~\ref{sec:Implementation}).
\subsection{One-step Optimization Objective for Joint Spectrum Estimation and SEMMD}
\label{sec-Problem Formulation}
\par The forward acquisition process of an X-ray CT scanner can be formulated using a nonlinear physical model~\cite{seo2012nonlinear}:
\begin{equation}
    \rho(\mathbf{r}) = -\ln\left[\int_\mathcal{E}\eta(E)\cdot\mathrm{e}^{-\int_\mathbf{r}\mu(\mathbf{x}, E)\mathrm{d}\mathbf{x}}\mathrm{d}E\right],\ \forall\mathbf{r}\in\mathbf{\Pi},
    \label{eq:nonlinear-model}
\end{equation}
where $\mathbf{\Pi}$ denotes the set of X-rays from a source, $\rho(\mathbf{r})$ is a projection by the X-ray $\mathbf{r}$ passing through the scanned object, $0\leq\eta(E)\leq1$ is the normalized energy spectrum that describes the distribution of the number of photons emitted by the X-ray source within the energy range $\mathcal{E}$, and $\mu(\mathbf{x}, E)$ is the linear attenuation coefficient (LAC) of observed object at position $\mathbf{x}$ for X-rays of energy $E$.
\par The LAC can further be represented as follows~\cite{jansen1980icru,beutel2000handbook}:
\begin{equation}
    \mu(\mathbf{x}, E) = \sigma(\mathbf{x})\cdot\kappa(\mathbf{x}, E),
    \label{eq:LAC-MAC}
\end{equation}
where $\sigma(\mathbf{x})$ denotes the density of the object at position $\mathbf{x}$, and $\kappa(\mathbf{x}, E)$ represents the mass attenuation coefficient (MAC) of the object at position $\mathbf{x}$ for X-rays of energy $E$.
\par In MMD theory~\cite{mendoncca2013flexible,xue2021multi,xue2020image,long2014multi}, the MAC of a mixture is often decomposed into a linear combination of the MACs of $M$ types of predefined basis materials, as shown below:
\begin{equation}
    \kappa(\mathbf{x}, E) = \sum_{i=1}^M\omega _i(\mathbf{x})\cdot\kappa_i(E),\ \
    \omega _i(\mathbf{x})=\frac{m_i(\mathbf{x})}{\sum_{j=1}^{M}m_j(\mathbf{x})},
    \label{eq:MAC分解}
\end{equation}
where $\kappa_i(E)$ is the MAC of $i$-th basis material for X-rays of energy $E$, $m_i(\mathbf{x})$ is the mass of the $i$-th basis material in the mixture at position $\mathbf{x}$, and $0\le\omega _i(\mathbf{x})\le1$ is the mass fraction of $i$-th basis material in the mixture at position $\mathbf{x}$.
\par Substituting Eq.~\eqref{eq:MAC分解} into Eq.~\eqref{eq:LAC-MAC}, we obtain:
\begin{equation}
    \mu(\mathbf{x}, E) = \sigma(\mathbf{x})\cdot\sum_{i=1}^M\frac{m_i(\mathbf{x})}{\sum_{j=1}^{M}m_j(\mathbf{x})}\cdot\kappa_i(E).
    \label{eq:MAC分解&LAC-MAC}
\end{equation}
\begin{figure}[t]
    \centering
    \includegraphics[width=\linewidth]{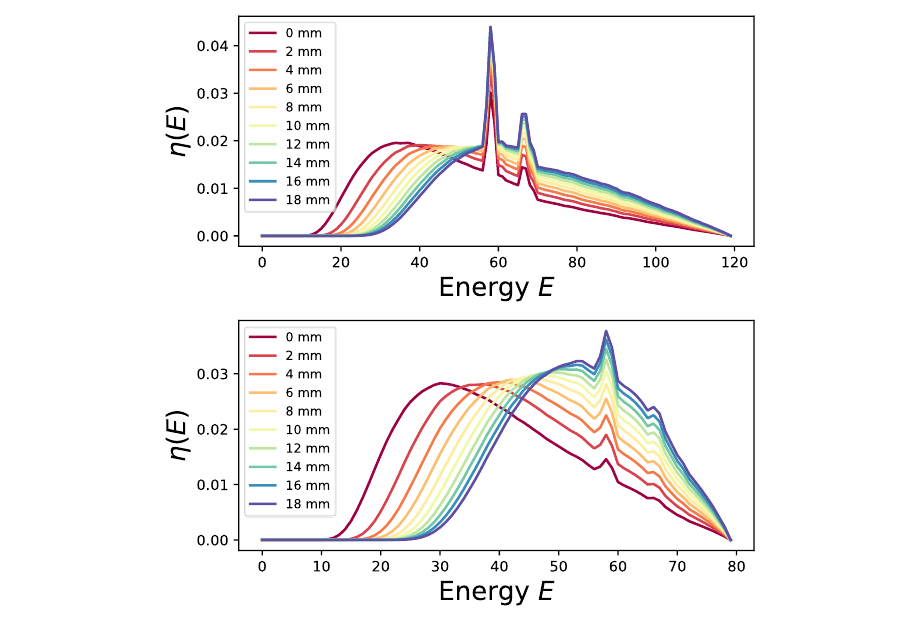}
    \caption{Two spectrum libraries with 10 different thickness Al tube filters at 120 kVP and 80 kVP generated by the SPEKTR toolkit~\cite{punnoose2016technical}.}
    \label{fig:fig_spectrum_lib}
\end{figure}
\par Based on the ideal solution assumption~\cite{mendoncca2010multi,long2014multi,mendoncca2013flexible}, the density of the mixture can be expressed as follows:
\begin{equation}
    \sigma(\mathbf{x}) = \frac{\sum_{i=1}^Mm_i(\mathbf{x})}{\sum_{i=1}^Mv_i(\mathbf{x})},
    \label{eq:ideal_solution}
\end{equation}
where $v_i(\mathbf{x})$ is the volume of the $i$-th basis material in the mixture at position $\mathbf{x}$. For $i$-th basis material, we have additionally:
\begin{equation}
    \kappa_i(E) = \frac{\mu_i(E)}{\sigma_i},\forall E,\ \ \sigma_i= \frac{m_i(\mathbf{x})}{v_i(\mathbf{x})},\ \forall\mathbf{x},
    \label{eq:材料基本性质}
\end{equation}
where $\mu_i(E)$ is the LAC of the $i$-th basis material for X-rays of energy $E$ and \roundtworevision{$\sigma_i$} is its density.
\par Substituting Eq.~\eqref{eq:ideal_solution} and Eq.~\eqref{eq:材料基本性质} into Eq.~\eqref{eq:MAC分解&LAC-MAC}, we get:
\begin{equation}
    \mu(\mathbf{x}, E) = \sum_{i=1}^M\alpha_i(\mathbf{x})\cdot\mu_i(E),\ \ \alpha_i(\mathbf{x})=\frac{v_i(\mathbf{x})}{\sum_{j=1}^Mv_j(\mathbf{x})},
    \label{eq:MMD}
\end{equation}
where $0\le\alpha_i(\mathbf{x})\le1$ is defined as the volume fraction of the $i$-th basis material in the mixture at position $\mathbf{x}$.
\par By combining Eq.~\eqref{eq:nonlinear-model} and Eq.~\eqref{eq:MMD}, we theoretically derive a physical forward model for SEMMD as follows:
\begin{equation}
        \mathcal{H}:\ \rho(\mathbf{r}) = -\ln\left[\int_\mathcal{E}\eta(E)\cdot\mathrm{e}^{-\sum\limits_{i=1}^M\mu_i(E)\cdot \int_\mathbf{r}\alpha_i(\mathbf{x})\mathrm{d}\mathbf{x}}\mathrm{d}E\right].
    \label{eq:MMD-model}
\end{equation}
\begin{figure*}[t]
    \centering
    \includegraphics[width=0.925\linewidth]{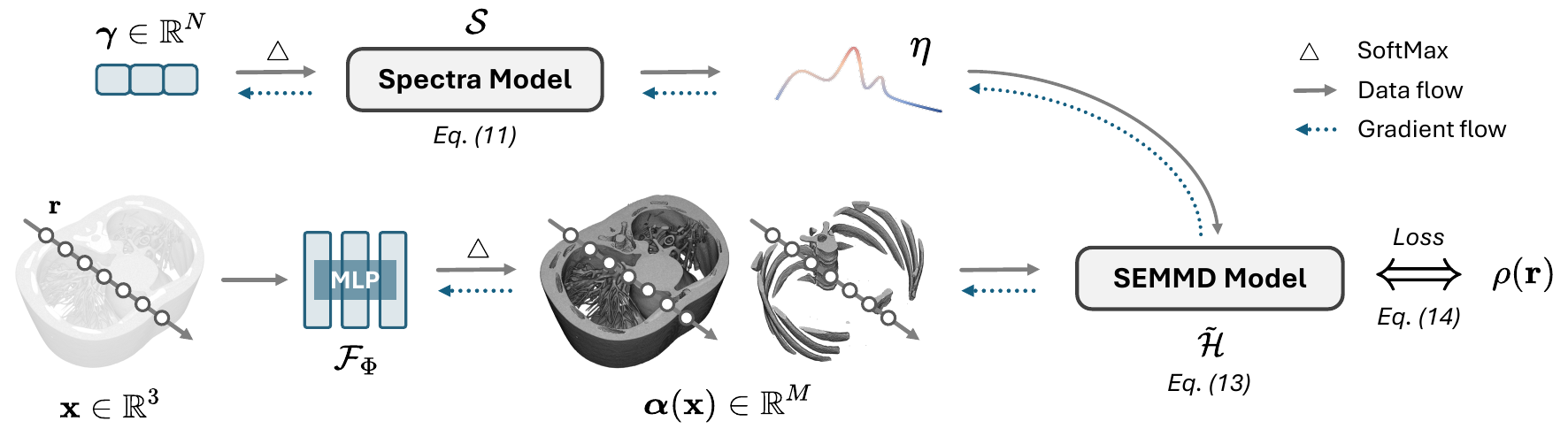}
    \caption{Optimization pipeline of the proposed \roundtworevision{JSolver}. Given SECT projection $\rho(\mathbf{r}), \forall \mathbf{r} \in \mathbf{\Pi}$, an MLP network $\mathcal{F}_\Phi$ maps multiple coordinates $\mathbf{x}$ along its X-ray $\mathbf{r}$ to the corresponding volume fractions $\boldsymbol{\alpha}(\mathbf{x}) = \mathcal{F}_{\Phi}(\mathbf{x})$. Concurrently, the spectrum $\eta$ is generated via a spectra model $\mathcal{S}$ (Eq.~\ref{eq:spectra_model}), controlled by learnable parameters $\boldsymbol{\gamma}$. The spectrum $\eta$ and the volume fractions $\boldsymbol{\alpha}(\mathbf{x}), \forall \mathbf{x} \in \mathbf{r}$ are then used to generate predicted SECT measurements $\hat{\rho}(\mathbf{r})$ via a discrete SEMMD model $\tilde{\mathcal{H}}$ (Eq.~\ref{eq:semmd}). Finally, the MLP network $\mathcal{F}_\Phi$ and spectrum parameters $\boldsymbol{\gamma}$ are optimized by minimizing a data consistency loss $\mathcal{L}_\text{DC}$ (Eq.~\ref{eq:loss-dc}), calculating the discrepancy between the predicted $\hat{\rho}(\mathbf{r})$ and the acquired SECT projections ${\rho}(\mathbf{r})$.}
    \label{fig:method}
\end{figure*}
\par In this work, we suppose that there are multiple predefined basis materials with the corresponding known LACs $\{\mu_i\}_{i=1}^M$. Then, our goal is to jointly estimate the energy spectrum $\eta(E),\forall E$ and reconstruct the volume fraction maps $\boldsymbol{\alpha}(\mathbf{x})=\{\alpha_i(\mathbf{x})\}_{i=1}^M,\forall\mathbf{x}$ directly from the SECT projections $\rho(\mathbf{r}),\forall\mathbf{r}\in\mathbf{\Pi}$. Mathematically, we establish our one-step optimization objective for joint spectrum estimation and SEMMD as follows:
\begin{equation}
    \begin{aligned}
    \eta^*,\ \boldsymbol{\alpha}^*=\arg\min_{\eta,\ \boldsymbol{\alpha}}\sum_{\mathbf{r}\in\mathbf{\Pi}}&\roundtworevision{\bigg\|\mathcal{H}(\eta,\ \boldsymbol{\alpha})-\rho(\mathbf{r})\bigg\|_1},\\
    \textrm{s. t.} \quad \int_\mathcal{E}\eta(E)\mathrm{d}E &= 1,\ \ \eta(E)\ge0,\forall E, \\
         \|\boldsymbol{\alpha}(\mathbf{x})\|_1&=1,\ \boldsymbol{\alpha}(\mathbf{x}) \succeq \mathbf{0},\forall\mathbf{x}.
    \end{aligned}
    \label{eq:objective}
\end{equation}  
where $\eta^*$ and $\boldsymbol{\alpha}^*$ denote the underlying optimal solutions.
\subsection{Representations for Spectrum and Volume Fractions}
\label{sec:Representations for Spectrum and Volume Fractions}
\par Due to the simultaneous estimation of multiple unknowns, the optimization problem, defined in Eq.~\eqref{eq:objective}, is highly under-determined, with numerous feasible solutions. We address this challenge by introducing two effective representations for the energy spectrum $\eta(E)$ and volume fraction maps $\boldsymbol{\alpha}(\mathbf{x})$. These representations can effectively constrain the solution space, significantly improving reconstruction quality.
\subsubsection{\textbf{\revision{Unconstrained Spectrum Estimation}}}
\par Previous studies on the X-ray energy spectrum estimation~\cite{zhao2014indirect,zhao2016segmentation,zhao2017segmentation,chang2019spectrum} often represent the unknown energy spectrum $\eta\in\mathbb{R}^L$ as a weighted summation of a set of predefined standard spectra $\{\eta_i\in\mathbb{R}^L\}_{i=1}^N$. As illustrated in Fig.~\ref{fig:fig_spectrum_lib}, these spectrum libraries can be easily generated using spectrum simulation tools, such as the SPEKTR toolkit~\cite{punnoose2016technical}. Formally, the spectra model can be written as follows:
\begin{equation}
    \begin{aligned}
            \eta(E)=&\sum_{i=1}^{N}\gamma_i\cdot\eta_i(E),\\\quad \textrm{s. t.}\quad &\sum_{i=1}^{N}\gamma_i=1,\gamma_i\ge0,\forall i,
    \end{aligned}
    \label{eq:spectra_model_prir}
\end{equation}
where $N$ is the number of predefined spectra, and $\gamma_i$ denotes the weight associated with the spectrum $\eta_i$. By leveraging this spectra model, the spectrum estimation task is reformulated as solving for the weights $\boldsymbol{\gamma}=\{\gamma_i\}_{i=1}^N$, which significantly alleviates the ill-posed nature of the inverse problem.
\par \revision{However, the spectrum model in Eq.~\eqref{eq:spectra_model_prir} requires optimization under non-negativity and sum-to-one constraints. We introduce a SoftMax transformation to optimize unconstrained parameters while satisfying both constraints exactly. Formally, our unconstrained spectrum model is defined as follows:}
\begin{equation}
    \begin{aligned}
    \mathcal{S}:\ \ \eta(E)=\sum_{i=1}^{N}&\text{SoftMax}(\gamma_i)\cdot\eta_i(E),\\
    \textrm{with} \quad&\text{SoftMax}(\gamma_i) = \frac{\mathrm{exp}{(\gamma_i)}}{\sum_{j=1}^{N}\mathrm{exp}{(\gamma_j)}},
\end{aligned}
    \label{eq:spectra_model}
\end{equation}
\revision{where the SoftMax transformation is differentiable, enabling gradient-based optimization of the parameters $\boldsymbol{\gamma}$. It strictly ensures non-negative spectrum weights that sum to one without explicit penalty terms.}
\subsubsection{\textbf{Volume Fraction Neural Representation}}
\par We propose representing the volume fraction maps as a continuous function of spatial coordinates $f$ as follows:
\begin{equation}
    f:\ \mathbf{x}\in\mathbb{R}^3 \rightarrow \boldsymbol{\alpha}(\mathbf{x})\in\mathbb{R}^M,
\end{equation}
where $\mathbf{x} = (x, y,z)$ denotes a coordinate in the 3D Cartesian system $\Omega = [-1, 1] \times [-1, 1] \times [-1, 1]$, and $\boldsymbol{\alpha}(\mathbf{x}) = [\alpha_1(\mathbf{x}), \dots, \alpha_M(\mathbf{x})]^\top$ represents the volume fractions at position $\mathbf{x}$. Note that our method is introduced in a 3D CT setting, but its application to 2D acquisition is straightforward.
\par The explicit expression of the volume fraction function $f$ is intractable. Instead, we learn its neural representation. Specifically, we use an MLP network $\mathcal{F}_\Phi$ that takes a 3D coordinate vector $\forall \mathbf{x} \in \Omega$ as input and outputs an $M$-dimensional vector corresponding to the volume fraction $\boldsymbol{\alpha}(\mathbf{x})$, to approximate the function (\ie $f \approx \mathcal{F}_\Phi: \mathbb{R}^3 \rightarrow \mathbb{R}^M$). Our key idea is to leverage the learning bias of neural networks toward low-frequency image patterns~\cite{rahaman2019spectral} as an implicit image prior, thereby eliminating sub-optimal volume fraction solutions in the optimization problem defined in Eq.~\eqref{eq:objective} and producing high-quality SEMMD reconstructions.
\subsection{Model Optimization}
\label{sec:optimization}
\par Fig.~\ref{fig:method} shows the workflow for optimizing the proposed \roundtworevision{JSolver} model. Given a SECT projection $\rho(\mathbf{r}), \forall \mathbf{r} \in \mathbf{\Pi}$, we first sample a set of coordinates $\mathbf{x}$ at a fixed interval $\Delta\mathbf{x}$ along the X-ray $\mathbf{r}$. These coordinates $\mathbf{x}$ are then fed into the MLP network $\mathcal{F}_{\Phi}$ to predict their corresponding volume fractions $\boldsymbol{\alpha}(\mathbf{x}), \forall \mathbf{x} \in \mathbf{r}$. Meanwhile, the energy spectrum $\eta$ is generated using the \revision{spectrum model} $\mathcal{S}$ (Eq.~\ref{eq:spectra_model}) from the spectrum library $\{\eta_i\}_{i=1}^N$ and the learnable \revision{spectrum parameters} $\boldsymbol{\gamma}$, initialized to one. Finally, we transform the estimated spectrum $\eta$ and the MLP-predicted volume fractions $\boldsymbol{\alpha}(\mathbf{x}) = \mathcal{F}_{\Phi}(\mathbf{x}), \forall \mathbf{x} \in \mathbf{r}$ into the SECT projections $\hat{\rho}(\mathbf{r})$ via a discrete form of the SEMMD forward model $\tilde{\mathcal{H}}$ (Eq.~\ref{eq:MMD-model}) as below:
\begin{equation}
    \begin{aligned}
        \tilde{\mathcal{H}}:\ \hat{\rho}(\mathbf{r}) = -\ln\left[\sum_{i=1}^{L}\eta(E_i)\cdot\mathrm{e}^{-\sum\limits_{j=1}^M\mu_j(E_i)\cdot \sum\limits_{\mathbf{x}\in\mathbf{r}}\alpha_j(\mathbf{x})\cdot\Delta\mathbf{x}}\right],
    \end{aligned}
    \label{eq:semmd}
\end{equation}
where $L$ is the length of the spectrum $\eta$.
\par Since the spectra model and the SEMMD forward model are differentiable with respect to the parameters $\boldsymbol{\gamma}$ and the volume fractions $\boldsymbol{\alpha}(\mathbf{x})$, we can jointly optimize the spectra parameters $\boldsymbol{\gamma}$ and the MLP network $\mathcal{F}_\Phi$ using gradient descent-based back-propagation algorithms to minimize a data consistency loss $\mathcal{L}_\text{DC}$ as follows:
\begin{equation}
    \roundtworevision{\mathcal{L}_\text{DC} = \frac{1}{|\mathbf{\Pi}|}\sum_{\mathbf{r}\in\mathbf{\Pi}}\|\hat{\rho}(\mathbf{r})-\rho(\mathbf{r})\|_1,}
    \label{eq:loss-dc}
\end{equation}
\roundtworevision{where $\mathbf{\Pi}$ denotes the full X-ray set.} \revision{The data consistency loss is evaluated in the log-projection domain. We use the $\ell_1$ loss because it is less sensitive than MSE to large residuals caused by low transmitted intensity, numerical clipping, or model mismatch. A Poisson negative log-likelihood is more directly suited to the raw photon-count domain and requires calibrated photon counts and incident intensity, which are not fully available for the real CT data used in this study. \roundtworevision{During stochastic optimization, we approximate the objective in Eq.~\eqref{eq:loss-dc} by averaging the $\ell_1$ residuals over a mini-batch $\mathbf{R}\subset\mathbf{\Pi}$. This subset is randomly resampled at each optimization step.}}
\subsection{Implementation Details}
\label{sec:Implementation}
\subsubsection{\textbf{Network Architecture}}
\par \revision{The MLP network $\mathcal{F}_{\Phi}$ consists of a multi-resolution hash encoding module~\cite{hash} and an MLP. The hash encoding maps low-dimensional spatial coordinates to high-dimensional feature vectors. Its hyper-parameters are set to $L=16$ levels, $F=8$ features per level, a hash-table size of $T=2^{18}$, a base resolution of $N_\text{min}=2$, and a per-level scale of $b=2$. The encoded coordinates are processed by two hidden FC layers, each containing 64 neurons followed by a ReLU activation. The output FC layer has $M$ channels and applies SoftMax to enforce the non-negativity and sum-to-one property of the volume fractions $\boldsymbol{\alpha}$.}
\subsubsection{\textbf{Optimization Hyper-parameters}}
\par \revision{\roundtworevision{JSolver} is implemented using PyTorch~\cite{paszke2019pytorch} and optimized using Adam~\cite{adam} with its default hyper-parameters. The learning rates of the INR and spectrum parameters are set to $1\times10^{-3}$ and $1\times10^{-2}$, respectively, and the total number of training epochs is 4,000. Each mini-batch contains 40 projection views, with two detector rays randomly sampled from each view, resulting in $|\mathbf{R}|=80$ rays per optimization step. The total training time is approximately 2 minutes on a single NVIDIA RTX 4070 Ti Super GPU with 16 GB memory. \roundtworevision{All training hyper-parameters remain consistent across the 2D datasets.} The exact configurations, training and evaluation code, and preprocessed simulated XCAT datasets are publicly available at \url{https://github.com/iwuqing/JSover}.}

\section{Experiments}
\par In this section, we evaluate the proposed \roundtworevision{JSolver}. First, we compare \roundtworevision{JSolver} with state-of-the-art SEMMD techniques on both simulated (Sec.~\ref{sec:exp_simulation}) and real-world (Sec.~\ref{sec:exp_real_solution} and Sec.~\ref{sec:exp_real_human}) CT data. Then, we investigate the impact of network architecture on \roundtworevision{JSolver} (Sec.~\ref{sec:exp_ab_network}). Finally, we assess its performance for undersampled CT (Sec.~\ref{sec:exp_undersampled}).
\subsection{Data Acquisition}
\subsubsection{\textbf{Simulated Digital XCAT Phantoms}}
\par
\begin{figure}[t]
    \centering
    \includegraphics[width=0.95\linewidth]{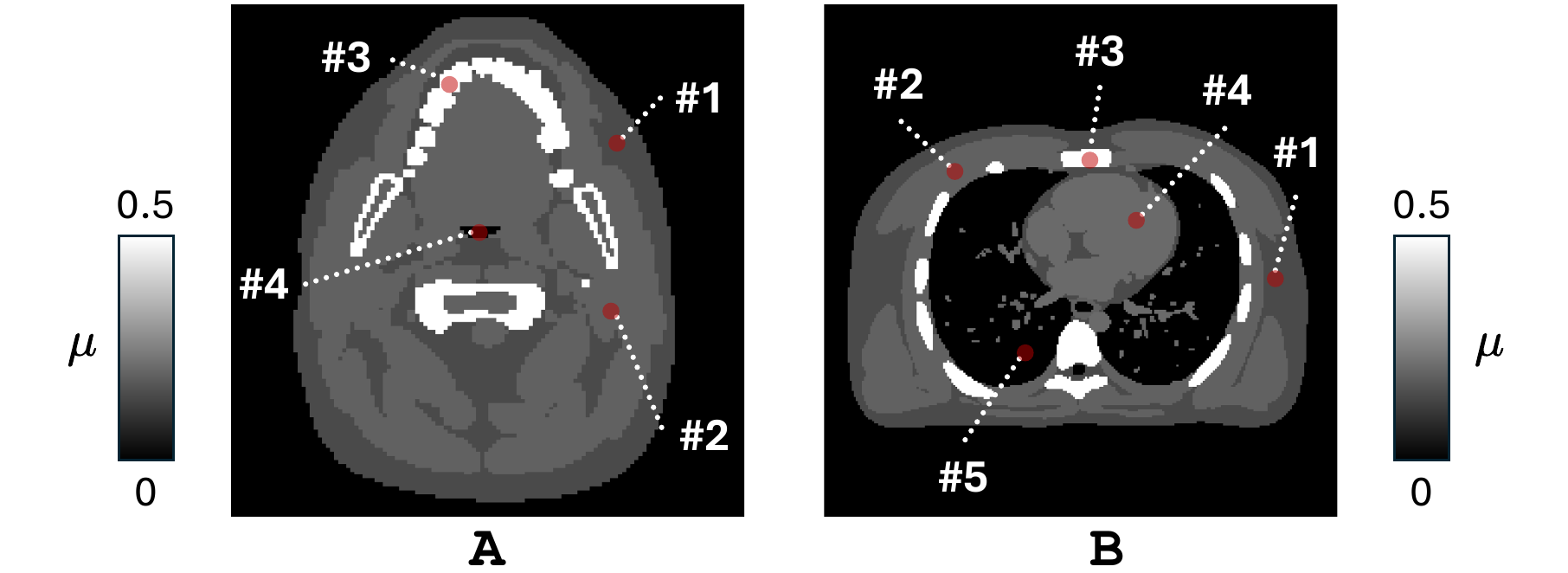}
    \caption{Two simulated digital XCAT phantoms. Here, the numbers indicate different regions of interest (ROI), listed in Table~\ref{tab:composition}.}
    \label{fig:digital_phantom_data}
\end{figure}

 We simulate two digital phantoms of 256$\times$256 size using the XCAT phantom program~\cite{segars20104d}, as shown in Fig.~\ref{fig:digital_phantom_data}, for quantitative evaluation. Table~\ref{tab:composition} shows their detailed material compositions. The LAC and density values of the basis materials are obtained from the NIST standard library~\cite{HubbellSeltzer2004} and the XCOM program~\cite{berger2009xcom}. We simulate two polychromatic X-ray sources with energy ranges of 120 and 80 kVP using the SPEKTR toolkit developed by Punnoose~\etal\cite{punnoose2016technical}. The corresponding spectra libraries are demonstrated in Fig.~\ref{fig:fig_spectrum_lib}. The CT measurements are generated with a size of 360$\times$363 by uniformly projecting 2D parallel-beam X-rays over angles in the range [0, 180$^\circ$). \revision{Based on the discrete polychromatic forward model in Eq.~\eqref{eq:semmd}, the noiseless projection $\hat{\rho}(\mathbf{r})$ is first computed for each ray $\mathbf{r}$. The detected photon count is sampled as $C(\mathbf{r})\sim\mathrm{Poisson}(I_0\exp[-\hat{\rho}(\mathbf{r})])$, followed by the logarithmic transformation $\rho(\mathbf{r})=-\ln[\max(C(\mathbf{r})/I_0,\epsilon)]$. We set $I_0=10^9$ and $\epsilon=10^{-10}$, corresponding to projection SNR values of 83.49 dB and 85.04 dB for phantoms A and B, respectively. Beam hardening is not added as a separate artifact. It arises naturally from the polychromatic spectrum and energy-dependent LACs, which preferentially attenuate low-energy photons.}
\begin{table}[t]
    \centering
    \caption{Material compositions of the two simulated digital XCAT phantoms, shown in Fig.~\ref{fig:digital_phantom_data}.}
    \tablestyle
    \begin{tabular}{cccccc}
    \toprule
    \multirow{2}{*}{\textbf{Phantom}}& \multirow{2}{*}{\textbf{ROI}} & \multicolumn{4}{c}{\textbf{Materials}} \\ \cmidrule(lr){3-6}
       & & \textbf{Adipose} & \textbf{Muscle} & \textbf{Bone} & \textbf{Air} \\ \midrule
      \multirow{4}{*}{\textbf{A}} & $\#$1 & 1 & 0 & 0 & 0 \\
                            & $\#$2 & 0.5 & 0.5 & 0  & 0 \\ 
                            & $\#$3 & 0 & 0 & 1  & 0 \\ 
                            & $\#$4 & 0 & 0 & 0 & 1 \\ \cmidrule{2-6} 
      \multirow{5}{*}{\textbf{B}} & $\#$1 & 1 & 0 & 0 & 0 \\
                            & $\#$2 & 0.5 & 0.5  & 0 & 0 \\ 
                            & $\#$3 & 0 & 0 & 1  & 0 \\ 
                            & $\#$4 & 0.3 & 0.7 & 0 & 0 \\ 
                            & $\#$5 & 0 & 0 & 0 & 1 \\ 
         \bottomrule
    \end{tabular}
    \label{tab:composition}
\end{table}

\begin{figure}[t]
    \centering
    \includegraphics[width=\linewidth]{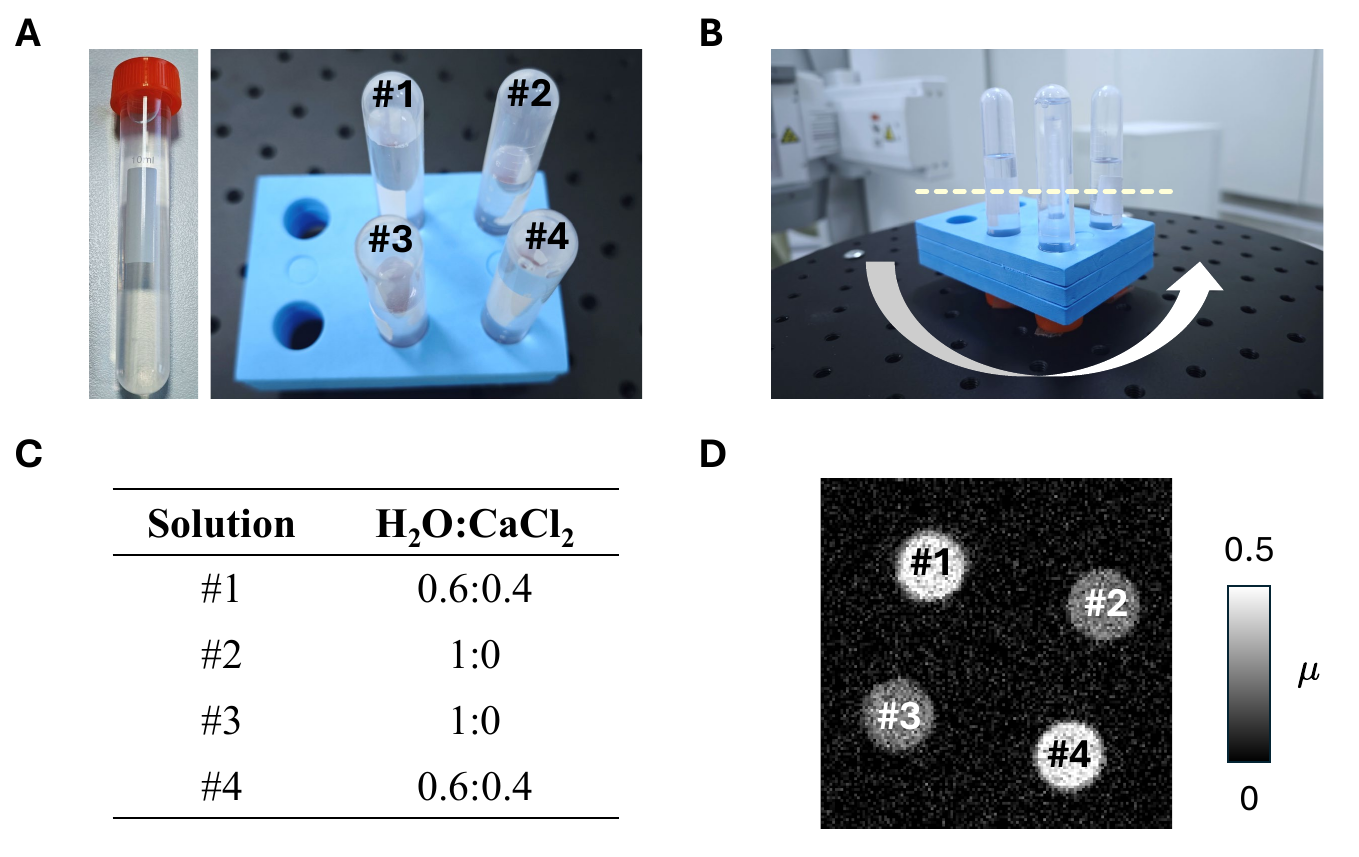}
    \caption{The experimental setup for the real-world solution phantoms: \textbf{A)} Four solution phantoms (10 ml), \textbf{B)} Material compositions of the four solutions, \textbf{C)} The cone-beam CT scanner in our lab, and \textbf{D)} The reconstructed SECT image of 128$\times$128 size using the FBP~\cite{fbp} algorithm.}
    \label{fig:real_setup}
\end{figure}
\begin{figure}[t]
    \centering
    \includegraphics[width=\linewidth]{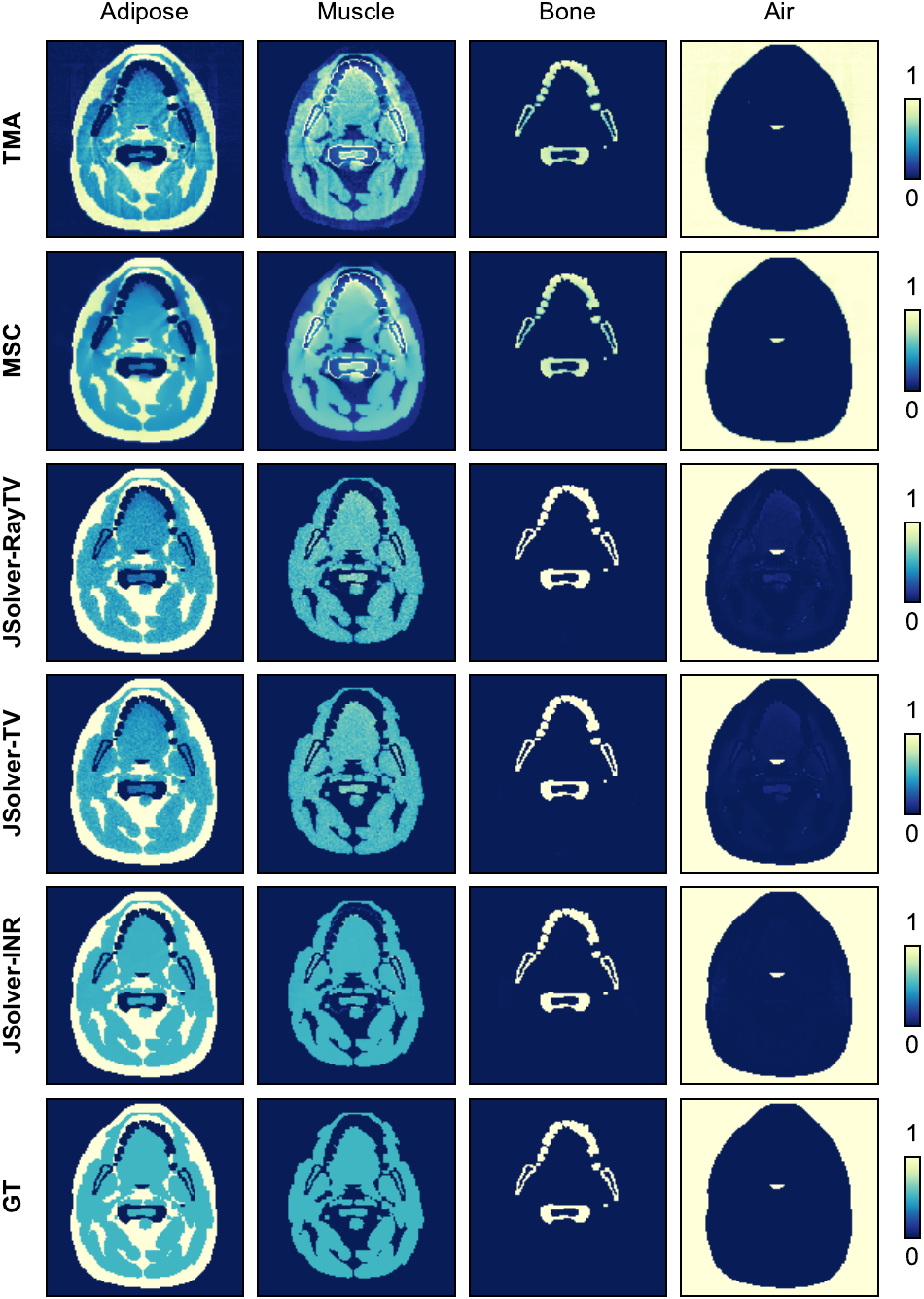}
    \caption{Qualitative comparison of SEMMD reconstructions by TMA~\cite{xue2020image}, MSC~\cite{xue2021multi}, \revision{\roundtworevision{JSolver}-RayTV, \roundtworevision{JSolver}-TV}, and \roundtworevision{JSolver}-INR on simulated XCAT phantom \textbf{\texttt{A}}.}
    \label{fig:simu_a}
\end{figure}
\subsubsection{\textbf{Real-world Solution Phantoms}}
\par To evaluate our method on real CT settings, we configure four solution phantoms: two containing only H$_2$O, and two consisting of H$_2$O and CaCl$_2$ mixed at a 0.6$:$0.4 ratio. As shown in Fig.~\ref{fig:real_setup}, we acquire data using a cone-beam CT scanner with the following protocols: image size, 128$\times$128; voxel size, 0.5$\times$0.5 mm$^2$; source voltage, 70 kVP; source current, 160 mA; Cu tube filter, 1.1 mm; source-to-isocenter distance, 750 mm; isocenter-to-detector distance, 750; and the number of views, 180.
\subsubsection{\textbf{Real-world Human Body Phantom}}
\par To test our method on clinical CT, we measure a 3D human body phantom using a commercial United Imaging Healthcare (UIH) uCT 768 scanner under a clinical helical CT protocol. The detailed acquisition parameters are as follows: image size, 512$\times$512$\times$175; voxel size, 0.5$\times$0.5$\times$0.6914 mm$^3$; source voltage, 120 kVP; source-to-isocenter distance, 570 mm; isocenter-to-detector distance, 490 mm; and number of views, 15,569.
\subsection{Evaluation Metrics}
\par For the reconstructed volume fractions $\boldsymbol{\alpha}$, following previous MMD works~\cite{xue2020image, xue2021multi, kim2022feasibility, long2014multi}, we report the mean and standard deviation (STD) of each region of interest (ROI), and compute the root-mean-square error (RMSE) as a quantitative metric, defined as 
\begin{equation}
    \mathrm{RMSE} = \sum_{i=1}^M\left(\sqrt{{\sum_{\mathbf{x}\in\Omega}(\alpha_i(\mathbf{x})-\hat{\alpha}_i(\mathbf{x}))^2}/{|\Omega|}}\right)/M
\end{equation}
where $\alpha_i$ and $\hat{\alpha}_i$ represent the GT and reconstructed volume fraction maps of the $i$-th basis material, respectively. 
\par \roundtworevision{For the estimated spectrum, we report the mean absolute error (MAE) between the GT and estimated normalized spectra.}
\begin{table*}[t]
    \centering
    \caption{Quantitative comparison of SEMMD reconstructions by TMA~\cite{xue2020image}, MSC~\cite{xue2021multi}, \revision{\roundtworevision{JSolver}-RayTV, \roundtworevision{JSolver}-TV}, and \roundtworevision{JSolver}-INR on the two simulated XCAT phantoms. \roundtworevision{\textbf{Bold} indicates ROI means closest to GT and minimum RMSE.}}
    \tablestyle
    \setlength{\tabcolsep}{4pt}
    \begin{tabular}{cccccccccc}
    \toprule
     \textbf{Phantom} & \textbf{Metric} & \textbf{ROI} & \textbf{Material} & \textbf{Truth} & \textbf{TMA}~\cite{xue2020image} & \textbf{MSC}~\cite{xue2021multi} & \revision{\textbf{\roundtworevision{JSolver}-RayTV}} & \revision{\textbf{\roundtworevision{JSolver}-TV}} & \textbf{\roundtworevision{JSolver}-INR} \\ \midrule
      \multirow{6}{*}{\textbf{A}} & \multirow{5}{*}{\textbf{Mean$\pm$STD}} & $\#$1 & Adipose & 1 & 0.8606$\pm$0.1387 & 0.8560$\pm$0.1360 & 0.9824$\pm$0.0412 & \revision{0.9871$\pm$0.0411} & \textbf{0.9912$\pm$0.0262} \\
                     & & \multirow{2}{*}{$\#$2} & Adipose & 0.5 & 0.3911$\pm$0.1147 & 0.3953$\pm$0.1057 & 0.4409$\pm$0.0712 & \revision{0.4522$\pm$0.0590} & \textbf{0.4883$\pm$0.0231} \\
                     & & & Muscle & 0.5 & 0.6089$\pm$0.1147 & 0.6047$\pm$0.1057 & 0.5335$\pm$0.0639 & \revision{0.5268$\pm$0.0510} & \textbf{0.5000$\pm$0.0226} \\
                     & & $\#$3 & Bone & 1 & 0.7531$\pm$0.1003 & 0.7501$\pm$0.0980 & 0.9999$\pm$0.0009 & \revision{\textbf{1.0000$\pm$0.0003}} & 0.9873$\pm$0.0228 \\
                     & & $\#$4 & Air & 1 & 0.9838$\pm$0.0370 & 0.9921$\pm$0.0365 & \textbf{1.0000$\pm$0.0003} & \revision{\textbf{1.0000$\pm$0.0001}} & 0.9996$\pm$0.0046 \\ \cmidrule{2-10}
                     & \textbf{RMSE} & - & - & - & 0.0981 & 0.0959 & 0.0363 & \revision{0.0247} & \textbf{0.0139} \\ \midrule
     \multirow{8}{*}{\textbf{B}} & \multirow{7}{*}{\textbf{Mean$\pm$STD}} & $\#$1 & Adipose & 1 & 0.8180$\pm$0.1557 & 0.8115$\pm$0.1524 & \textbf{0.9870$\pm$0.0288} & \revision{0.9835$\pm$0.0592} & 0.9792$\pm$0.0588 \\
                     & & \multirow{2}{*}{$\#$2} & Adipose & 0.5 & 0.3820$\pm$0.1721 & 0.3880$\pm$0.1582 & 0.4793$\pm$0.0566 & \revision{0.4850$\pm$0.0532} & \textbf{0.4854$\pm$0.0555} \\
                     & & & Muscle & 0.5 & 0.6178$\pm$0.1730 & 0.6119$\pm$0.1591 & 0.5103$\pm$0.0535 & \revision{0.5072$\pm$0.0457} & \textbf{0.5056$\pm$0.0548} \\
                     & & $\#$3 & Bone & 1 & 0.8602$\pm$0.1040 & 0.8592$\pm$0.1029 & \textbf{1.0000$\pm$0.0004} & \revision{\textbf{1.0000$\pm$0.0003}} & 0.9859$\pm$0.0269 \\
                     & & \multirow{2}{*}{$\#$4} & Adipose & 0.3 & 0.2649$\pm$0.1762 & \roundtworevision{\textbf{0.2712$\pm$0.1596}} & 0.2187$\pm$0.0550 & \revision{0.2149$\pm$0.0613} & \roundtworevision{0.3431$\pm$0.0541} \\
                     & & & Muscle & 0.7 & 0.7347$\pm$0.1776 & \roundtworevision{\textbf{0.7284$\pm$0.1613}} & \roundtworevision{0.7485$\pm$0.0480} & \revision{0.7505$\pm$0.0453} & 0.6449$\pm$0.0533 \\
                     & & $\#$5 & Air & 1 & 0.9574$\pm$0.0877 & 0.9647$\pm$0.0876 & \textbf{0.9999$\pm$0.0013} & \revision{\textbf{0.9999$\pm$0.0010}} & 0.9969$\pm$0.0214 \\ \cmidrule{2-10}
                     & \textbf{RMSE} & - & - & - & 0.1027 & 0.0992 & 0.0223 & \revision{\textbf{0.0198}} & 0.0272 \\
         \bottomrule
    \end{tabular}
    \label{tab:digital_phantom_data}
\end{table*}

\begin{figure}[!t]
    \centering
    \includegraphics[width=\linewidth]{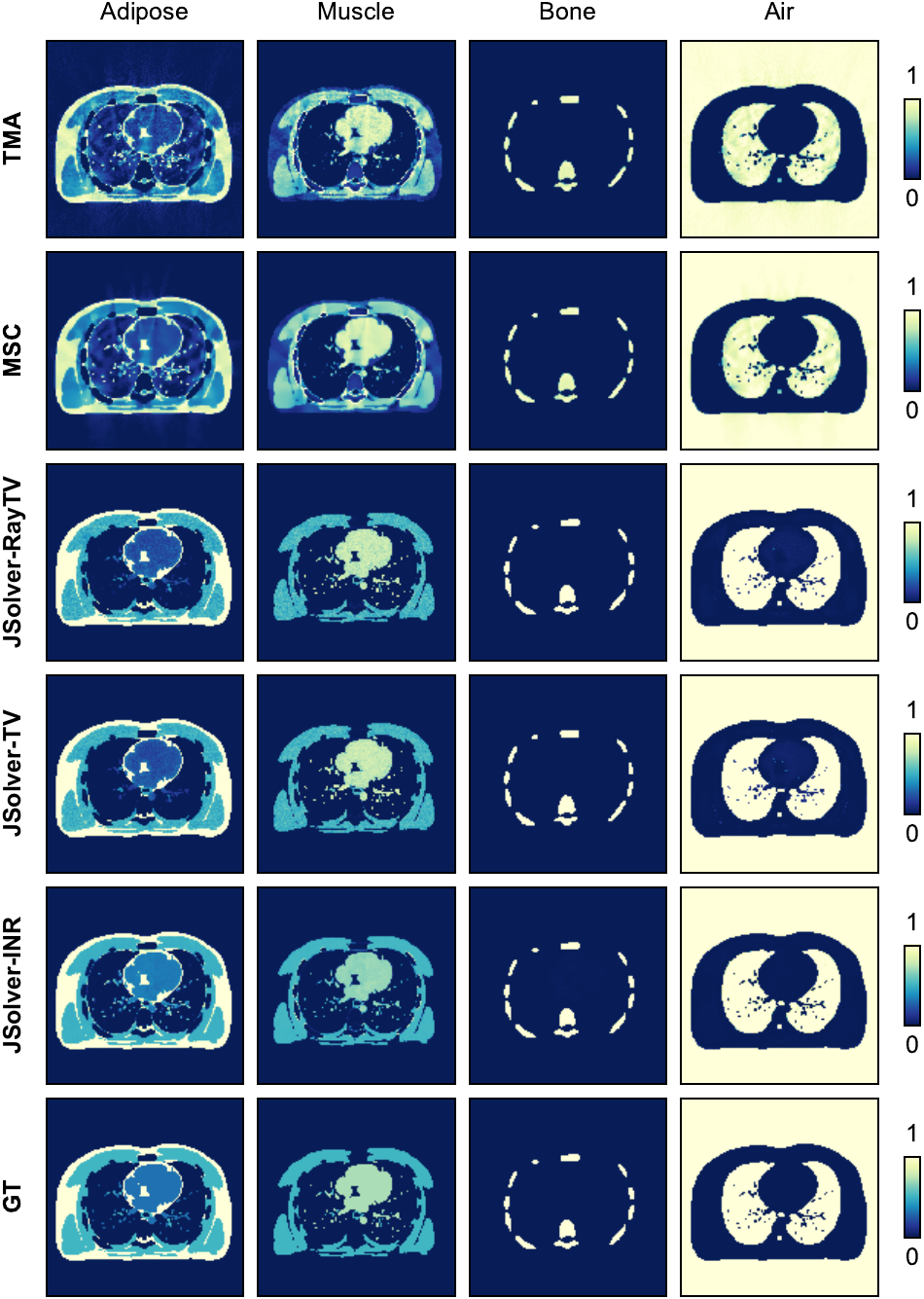}
    \caption{Qualitative comparison of SEMMD reconstructions by TMA~\cite{xue2020image}, MSC~\cite{xue2021multi}, \revision{\roundtworevision{JSolver}-RayTV, \roundtworevision{JSolver}-TV}, and \roundtworevision{JSolver}-INR on simulated XCAT phantom \textbf{\texttt{B}}.}
    \label{fig:simu_b}
\end{figure}
\subsection{Compared Methods}
\par We first \revision{employ} TMA~\cite{xue2020image} and MSC~\cite{xue2021multi}, two representative optimization-based SEMMD approaches, as baseline comparisons. Both models are implemented according to their original papers, with hyperparameters carefully tuned to ensure a fair comparison. To evaluate the contribution of the proposed joint optimization framework and the effectiveness of INR, we compare \roundtworevision{JSolver}-INR with two variants using different TV regularizers. Both variants directly optimize discrete material volume fraction maps under the same polychromatic forward model and spectrum estimation model. Their general optimization problem is
\begin{equation}
    \arg\min_{\boldsymbol{\gamma},\ \boldsymbol{\alpha}}
    \roundtworevision{\frac{1}{|\mathbf{\Pi}|}}
    \sum_{\mathbf{r}\in\mathbf{\Pi}}
    \left\|\tilde{\mathcal{H}}\left(\mathcal{S}(\boldsymbol{\gamma}),\boldsymbol{\alpha}\right)-\rho(\mathbf{r})\right\|_1
    +\lambda \mathcal{L}_\mathrm{reg}(\boldsymbol{\alpha}),
    \label{eq:tv}
\end{equation}
where $\mathcal{L}_\mathrm{reg}$ denotes the regularization term and $\lambda$ denotes its weight. \roundtworevision{The data consistency term is averaged over the full ray set $\mathbf{\Pi}$, as in Eq.~\eqref{eq:loss-dc}.}

\par The first variant uses a ray-wise finite-difference regularizer:
\begin{equation}
    \mathcal{L}_\mathrm{RayTV}(\boldsymbol{\alpha})=
    \roundtworevision{\frac{1}{M|\mathbf{\Pi}|}\sum_{\mathbf{r}\in\mathbf{\Pi}}}
    \sum_{\mathbf{x}_a,\mathbf{x}_b\in\mathbf{r}}
    \left\|\boldsymbol{\alpha}(\mathbf{x}_a)-\boldsymbol{\alpha}(\mathbf{x}_b)\right\|_1,
    \label{eq:ray_tv}
\end{equation}
\roundtworevision{where the inner sum includes only adjacent sampling coordinates $\mathbf{x}_a$ and $\mathbf{x}_b$ on X-ray $\mathbf{r}$. The regularizer is averaged over rays and basis materials. At each optimization step, both the data consistency term and the ray-wise regularizer are approximated by their averages over the randomly sampled mini-batch $\mathbf{R}\subset\mathbf{\Pi}$.} Since this term is not classical image-domain TV, we denote this variant as \roundtworevision{JSolver}-RayTV, with $\mathcal{L}_\mathrm{reg}=\mathcal{L}_\mathrm{RayTV}$ and $\lambda=\lambda_\mathrm{R}$ in Eq.~\eqref{eq:tv}.

\par The second variant uses classical anisotropic image-domain TV, which penalizes the horizontal and vertical finite differences of the complete material maps:
\begin{equation}
    \mathcal{L}_\mathrm{TV}(\boldsymbol{\alpha})=
    \roundtworevision{\frac{1}{M}\sum_{m=1}^{M}\left(\frac{\left\|D_x\alpha_m\right\|_1}{N_x}+\frac{\left\|D_y\alpha_m\right\|_1}{N_y}\right)},
    \label{eq:classical_tv}
\end{equation}
where $D_x$ and $D_y$ denote the horizontal and vertical finite-difference operators, respectively. \roundtworevision{$N_x$ and $N_y$ are the numbers of horizontal and vertical finite-difference entries in each material map. Thus, each directional term is averaged over its entries, consistent with the implementation. Unlike the ray-wise regularizer, this term is evaluated on the complete material maps at every optimization step.} We denote this variant as \roundtworevision{JSolver}-TV, with $\mathcal{L}_\mathrm{reg}=\mathcal{L}_\mathrm{TV}$ and $\lambda=\lambda_\mathrm{TV}$ in Eq.~\eqref{eq:tv}. We set $\lambda_\mathrm{R}=3\times10^{-4}$ for \roundtworevision{JSolver}-RayTV and $\lambda_\mathrm{TV}=0.1$ for \roundtworevision{JSolver}-TV. The two variants use Adam~\cite{adam} with a learning rate of $1\times10^{-2}$, and both variants are optimized for 8,000 iterations.
\begin{figure}[t]
    \centering
    \includegraphics[width=\linewidth]{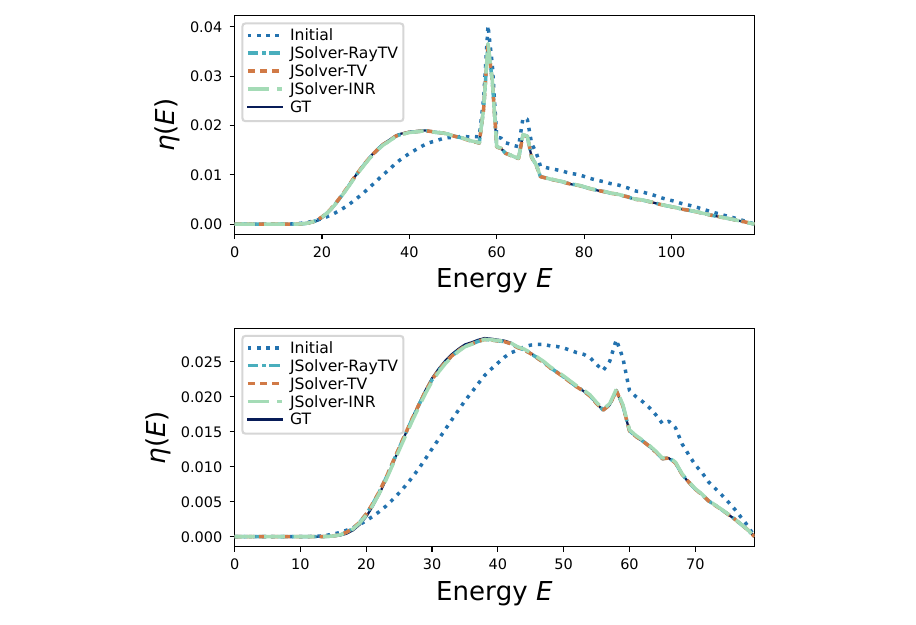}
    \caption{Comparison of the initial, \revision{\roundtworevision{JSolver}-RayTV, \roundtworevision{JSolver}-TV}, \roundtworevision{JSolver}-INR, and GT spectra on the simulated XCAT phantoms \textbf{\texttt{A}} (top) and \textbf{\texttt{B}} (bottom). Here, ``initial'' refers to the average of the pre-defined library, given by $\sum_{i=1}^N \eta_i/N$.}
    \label{fig:fig_spectrum_pre}
\end{figure}
\subsection{Results on Simulated Digital XCAT Phantoms}
\label{sec:exp_simulation}
\subsubsection{\textbf{\roundtworevision{Comparison with Existing Methods}}}
\par Fig.~\ref{fig:simu_a} and Fig.~\ref{fig:simu_b} show the qualitative \revision{results}. Visually, TMA~\cite{xue2020image} exhibits severe X-ray beam hardening artifacts, since it performs SEMMD on \roundtworevision{FBP SECT reconstructions}~\cite{fbp}. In comparison, MSC~\cite{xue2021multi} effectively reduces these artifacts by adding TV regularization and material sparsity constraints, but its reconstruction accuracy is still limited. It is also observed that both image-domain optimization methods tend to give better decomposition results when the basis materials have clearly different MAC profiles. For example, the decomposed maps for bone and air match the GT better than those for adipose and muscle. 
\par \revision{In contrast, \roundtworevision{JSolver}-RayTV, \roundtworevision{JSolver}-TV, and \roundtworevision{JSolver}-INR avoid beam hardening artifacts through projection-domain optimization. \roundtworevision{JSolver}-TV produces cleaner and more spatially coherent material maps than \roundtworevision{JSolver}-RayTV. \roundtworevision{JSolver-INR preserves the overall structure and local details through its continuous representation.}}
\par \revision{The quantitative results are presented in Table~\ref{tab:digital_phantom_data}. \roundtworevision{JSolver}-TV achieves lower material RMSE than \roundtworevision{JSolver}-RayTV on both phantoms. \roundtworevision{On phantom B, JSolver-TV achieves a lower material RMSE than JSolver-INR (0.0198 vs. 0.0272), whereas JSolver-INR obtains the lowest RMSE on phantom A (0.0139). This difference is consistent with TV favoring piecewise-constant regions, while INR estimates several mixed-material fractions more accurately on phantom B (Table~\ref{tab:digital_phantom_data}), indicating complementary strengths of the two priors.} All three \roundtworevision{JSolver} variants substantially outperform TMA~\cite{xue2020image} and MSC~\cite{xue2021multi}.}
\begin{figure}[t]
    \centering
    \includegraphics[width=\linewidth]{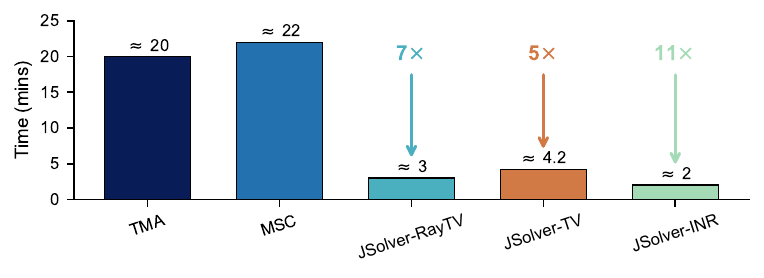}
    \caption{Comparison of reconstruction speed among TMA~\cite{xue2020image}, MSC~\cite{xue2021multi}, \revision{\roundtworevision{JSolver}-RayTV, \roundtworevision{JSolver}-TV}, and \roundtworevision{JSolver}-INR on the two simulated XCAT phantoms.}
    \label{fig:fig_time}
\end{figure}
\par \revision{The estimated spectra for the two simulated XCAT phantoms are presented in Fig.~\ref{fig:fig_spectrum_pre}. TMA~\cite{xue2020image} and MSC~\cite{xue2021multi} are excluded because they do not estimate the X-ray spectrum. The initial spectrum is the average of the pre-defined library, computed as $\sum_{i=1}^N \eta_i/N$. \roundtworevision{JSolver}-RayTV, \roundtworevision{JSolver}-TV, and \roundtworevision{JSolver}-INR all produce spectrum estimates that closely match the GT curves.}
\par Fig.~\ref{fig:fig_time} compares the mean reconstruction times \roundtworevision{on the two XCAT phantoms}. TMA~\cite{xue2020image} performs image-domain optimization in a \textit{voxel-wise} manner. As a result, it needs to solve 65,536 independent optimization problems at most for decomposed maps of size 256$\times$256, requiring approximately 20 minutes on a single NVIDIA \roundtworevision{RTX 4070 Ti Super} GPU. MSC~\cite{xue2021multi}, which is built upon TMA, takes even longer, with an average runtime of around 22 minutes. In contrast, our \roundtworevision{JSolver} models directly solve the SEMMD problem from SECT measurements. This process is comparable to conventional CT reconstruction in terms of computation. \revision{On the same hardware, \roundtworevision{JSolver}-RayTV, \roundtworevision{JSolver}-TV, and \roundtworevision{JSolver}-INR require approximately 3, 4.2, and 2 minutes, respectively, demonstrating significantly improved computational efficiency.}
\subsubsection{\textbf{\roundtworevision{Ablation Study on Spectrum Estimation}}}
\par \roundtworevision{To evaluate the contribution of spectrum estimation, we compare three spectrum settings on the two 256$\times$256 XCAT phantoms. Fixed initial uses the average of the pre-defined spectrum library throughout optimization. Learned jointly estimates the spectrum and material maps, while Fixed GT uses the ground-truth spectrum as a reference. All other settings remain unchanged.}
\par
\begin{figure}[t]
    \centering
    \includegraphics[width=\linewidth]{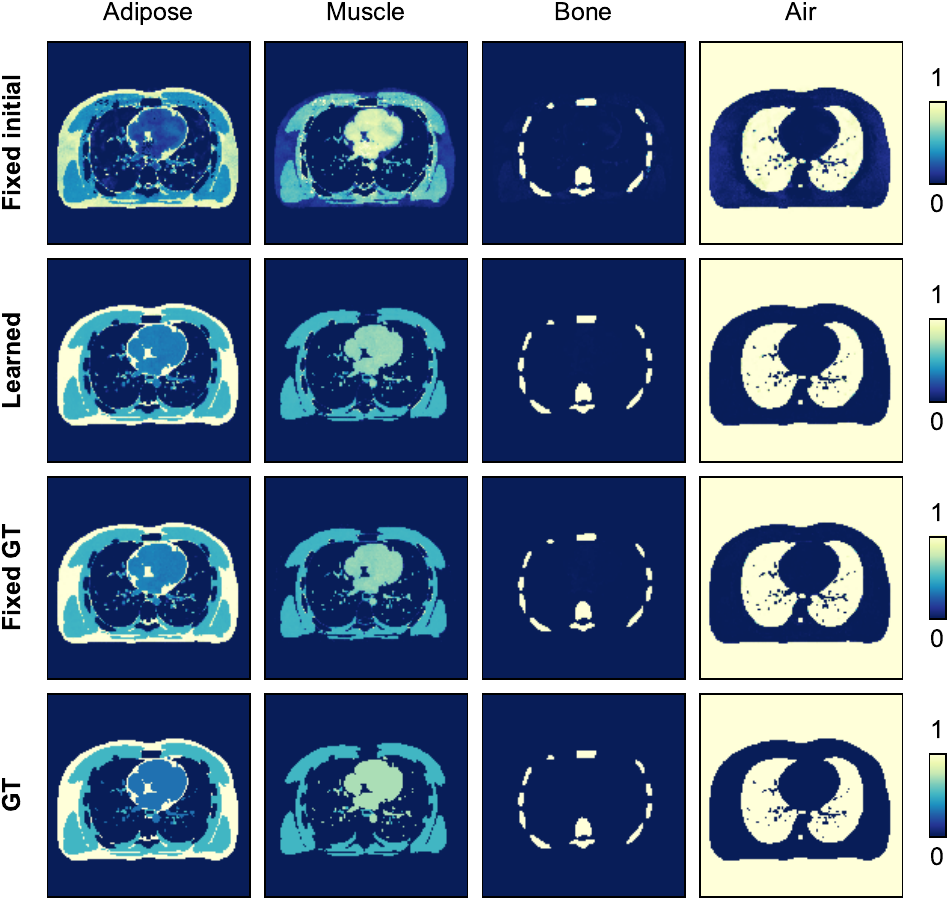}
    \caption{\roundtworevision{Qualitative comparison of JSolver-INR using learned and fixed spectra on the simulated XCAT phantom \textbf{\texttt{B}}. ``Fixed initial'' denotes the average of the pre-defined spectrum library, while ``Fixed GT'' denotes the ground-truth spectrum. The last row shows the GT material maps. All material maps use the same display range of [0, 1].}}
    \label{fig:fixed_spectrum}
\end{figure}

 \roundtworevision{Table~\ref{tab:fixed_spectrum} and Fig.~\ref{fig:fixed_spectrum} present the quantitative and qualitative results. Compared with the fixed initial spectrum, joint spectrum estimation reduces material RMSE from 0.0785 to 0.0139 on phantom A and from 0.0773 to 0.0272 on phantom B. The learned spectrum also improves the separation of adipose and muscle and produces material maps close to those obtained with the GT spectrum. These results show that updating the spectrum is important for correcting an inaccurate initial spectrum within the proposed joint framework.}
\begin{table}[t]
    \centering
    \caption{\roundtworevision{Quantitative comparison of JSolver-INR using learned and fixed spectra on the two simulated XCAT phantoms. ``Fixed initial'' denotes the average of the pre-defined spectrum library, while ``Fixed GT'' denotes the ground-truth spectrum. The best performances are highlighted in bold.}}
    \label{tab:fixed_spectrum}
    \begingroup
    \color{RevisionRed}
    \ifdefined\CleanVersion\color{black}\fi
    \tablestyle
    \begin{tabular}{cccc}
        \toprule
        \multirow{2}{*}{\textbf{Phantom}} & \multirow{2}{*}{\textbf{Spectrum Setting}} & \textbf{SEMMD} & \textbf{Spectrum} \\
        \cmidrule(lr){3-4}
        & & \textbf{RMSE} & \textbf{MAE} \\
        \midrule
        \multirow{3}{*}{\textbf{A}} & Fixed initial & 0.0785 & 0.001844 \\
          & Learned & \textbf{0.0139} & 0.000061 \\
          & Fixed GT & 0.0143 & \textbf{0.000000} \\
        \midrule
        \multirow{3}{*}{\textbf{B}} & Fixed initial & 0.0773 & 0.003536 \\
          & Learned & 0.0272 & 0.000136 \\
          & Fixed GT & \textbf{0.0248} & \textbf{0.000000} \\
        \bottomrule
    \end{tabular}
    \endgroup
\end{table}

\subsubsection{\textbf{\roundtworevision{Robustness to Out-of-Library Spectra}}}
\par
\begin{figure}[t]
    \centering
    \includegraphics[width=\linewidth]{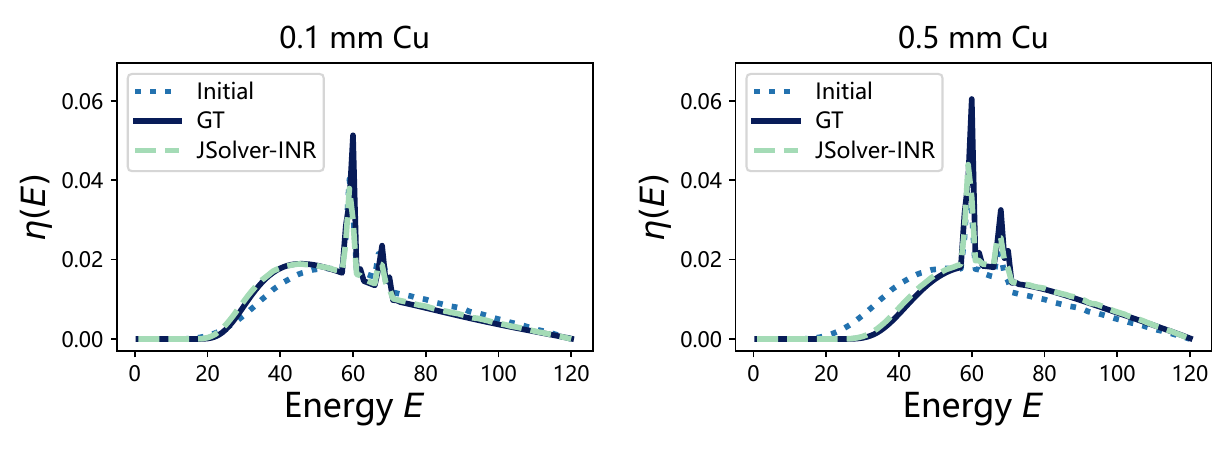}
    \vspace{1pt}
    \includegraphics[width=\linewidth]{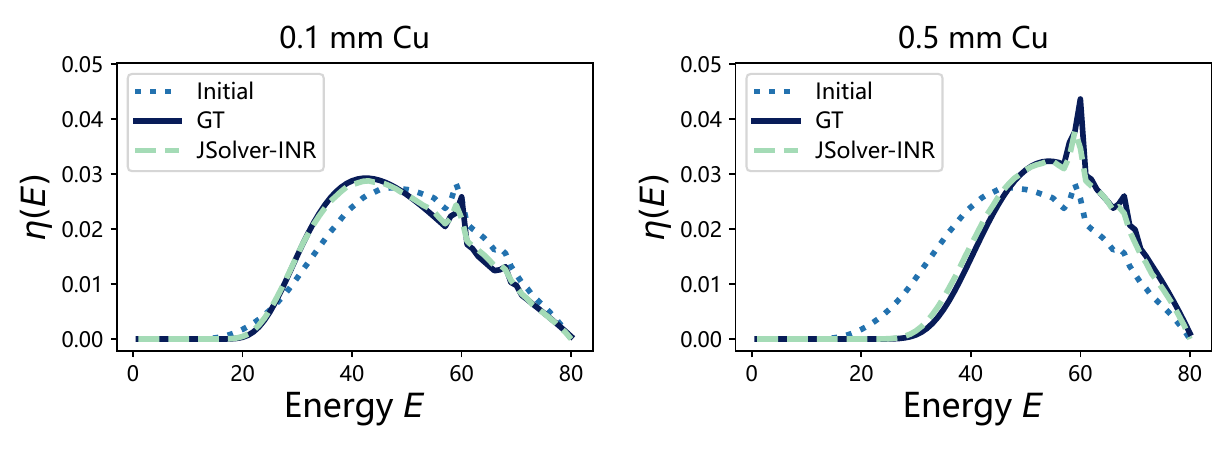}
    \caption{\revision{Comparison of the initial, \roundtworevision{JSolver}-INR, and GT spectra for the out-of-library Cu-filtered experiments on simulated XCAT phantoms \textbf{\texttt{A}} (top) and \textbf{\texttt{B}} (bottom). The target spectra use 0.1 mm and 0.5 mm Cu filters, while reconstruction uses the $N=10$ Al-filtered spectrum library.}}
    \label{fig:out_of_library}
\end{figure}

 \revision{We further evaluate robustness when the target spectrum is not represented by the pre-defined library. The target spectra are generated using 0.1 mm and 0.5 mm Cu filters, while \roundtworevision{JSolver}-INR retains the original $N=10$ library containing only Al-filtered spectra. Figure~\ref{fig:out_of_library} and Table~\ref{tab:out_of_library} present the spectrum comparisons and quantitative results, respectively. Compared with the in-library setting, the spectrum recovery error increases under the Cu filtration mismatch, particularly for 0.5 mm Cu. Nevertheless, the estimated spectra preserve the main spectral shapes, and the material RMSE remains between 0.0249 and 0.0305. These results show that \roundtworevision{JSolver}-INR retains acceptable decomposition accuracy under spectrum-library mismatch, while sufficient library coverage remains important for high-precision spectrum recovery.}
\begin{table}[t]
    \centering
    \caption{\revision{Quantitative results of \roundtworevision{JSolver}-INR for out-of-library Cu-filtered spectra using the $N=10$ Al-filtered spectrum library on the two simulated XCAT phantoms.}}
    \begingroup
    \tablestyle
    \begin{tabular}{cccc}
        \toprule
        \multirow{2}{*}{\textbf{Phantom}} & \multirow{2}{*}{\shortstack{\textbf{Cu Filter}\\\textbf{(mm)}}} & \textbf{\roundtworevision{SEMMD}} & \textbf{Spectrum} \\
        \cmidrule(lr){3-4}
        & & \textbf{RMSE} & \textbf{MAE} \\
        \midrule
        \multirow{2}{*}{\textbf{A}} & 0.1 & 0.0253 & 0.000616 \\
        & 0.5 & 0.0249 & 0.000749 \\
        \midrule
        \multirow{2}{*}{\textbf{B}} & 0.1 & 0.0281 & 0.000376 \\
        & 0.5 & 0.0305 & 0.000846 \\
        \bottomrule
    \end{tabular}
    \endgroup
    \label{tab:out_of_library}
\end{table}

\begin{figure}[t]
    \centering
    \includegraphics[width=0.825\linewidth]{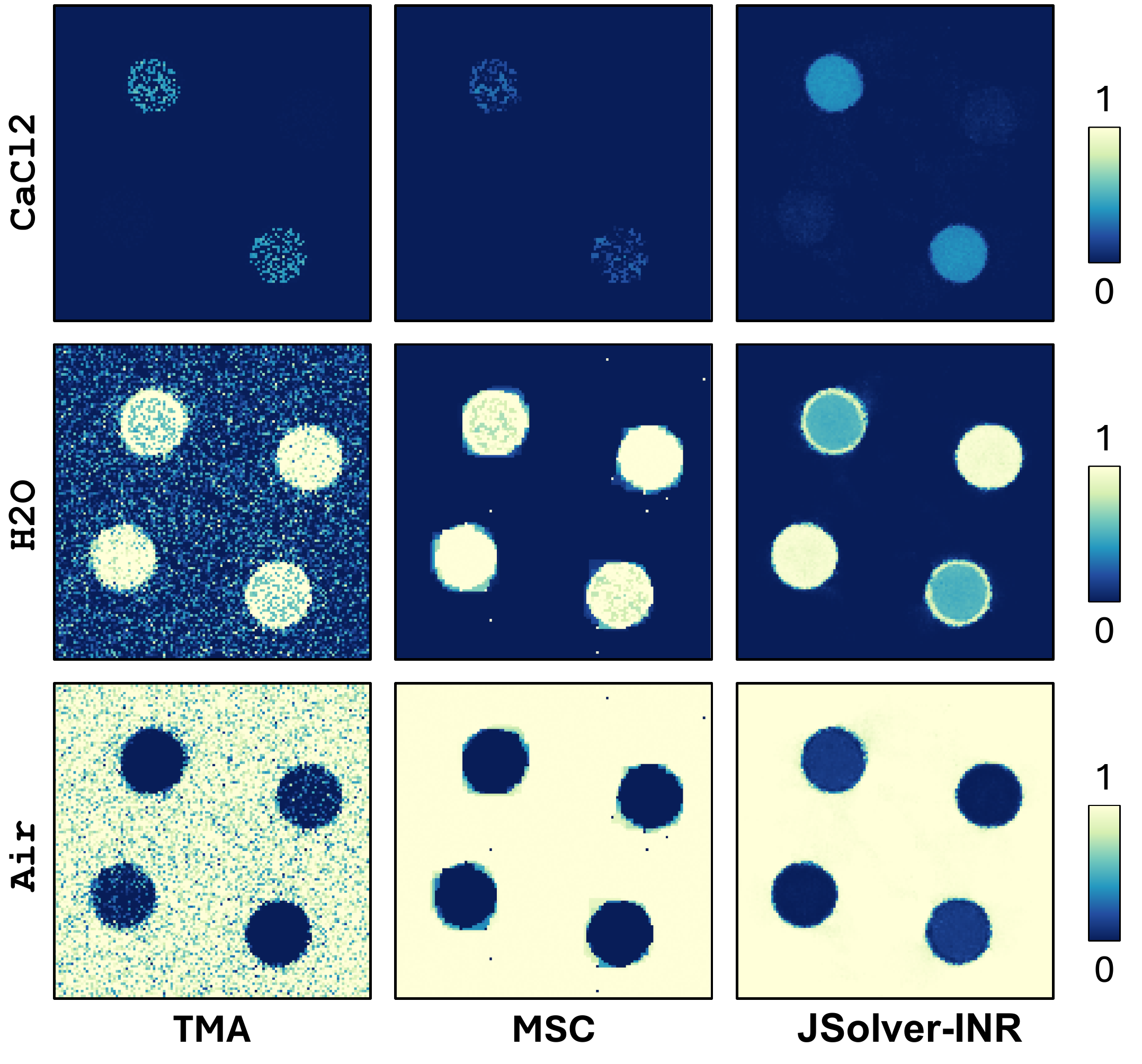}
    \caption{Qualitative comparison of SEMMD reconstructions by TMA~\cite{xue2020image}, MSC~\cite{xue2021multi}, our \roundtworevision{JSolver}-INR on the real-world solution phantoms.}
    \label{fig:fig_real_solution}
\end{figure}
\begin{table}[t]
    \centering
    \caption{Quantitative comparison of SEMMD reconstructions by TMA~\cite{xue2020image}, MSC~\cite{xue2021multi}, and our \roundtworevision{JSolver}-INR on the real-world solution phantoms. The best performances are highlighted in \textbf{bold}.}
    \tablestyle
    \begin{tabular}{cccccc}
    \toprule
     \textbf{Solution} & \textbf{Material} & \textbf{Truth}&\textbf{TMA}~\cite{xue2020image}& \textbf{MSC}~\cite{xue2021multi}&\textbf{\roundtworevision{JSolver}-INR} \\ \midrule
     \multirow{2}{*}{$\#$1} & CaCl$_2$& 0.4 & 0.149$\pm$0.205&0.074$\pm$0.108&\textbf{0.341$\pm$0.0857} \\
      & H$_2$O & 0.6 & 0.848$\pm$0.206&0.923$\pm$0.109&\textbf{0.543$\pm$0.0968} \\  \cmidrule{2-6}
     $\#$2 & H$_2$O & 1 & 0.932$\pm$0.141&\textbf{0.999$\pm$0.016}&0.952$\pm$0.0200 \\ \cmidrule{2-6}
     $\#$3 & H$_2$O & 1 & 0.918$\pm$0.150&\textbf{0.986$\pm$0.073}&0.949$\pm$0.0200 \\  \cmidrule{2-6}
      \multirow{2}{*}{$\#$4} & CaCl$_2$& 0.4 & 0.142$\pm$0.201&0.061$\pm$0.093&\textbf{0.340$\pm$0.0682} \\
      & H$_2$O & 0.6 & 0.840$\pm$0.217&0.924$\pm$0.121&\textbf{0.538$\pm$0.0808} \\
    \bottomrule
    \end{tabular}
    \label{tab:real_solution}
\end{table}

\begin{figure}[t]
    \centering
    \includegraphics[width=\linewidth]{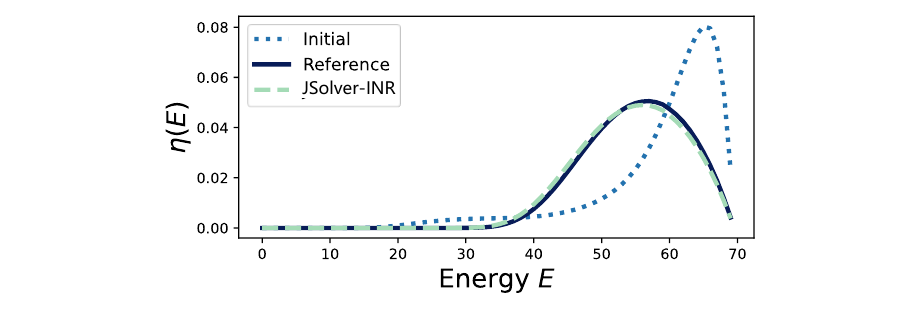}
    \caption{Comparison of the initial, \roundtworevision{JSolver}-INR, and reference spectra on the real-world solution phantom. Here, ``initial" refers to the average of the pre-defined library, given by $\sum_{i=1}^N \eta_i/N$, while ``reference'' is generated by the SPEKTR toolkit~\cite{punnoose2016technical} with known X-ray source configurations.}
    \label{fig:fig_spectrum_pre_solution} 
\end{figure}
\begin{figure}[t]
    \centering
    \includegraphics[width=\linewidth]{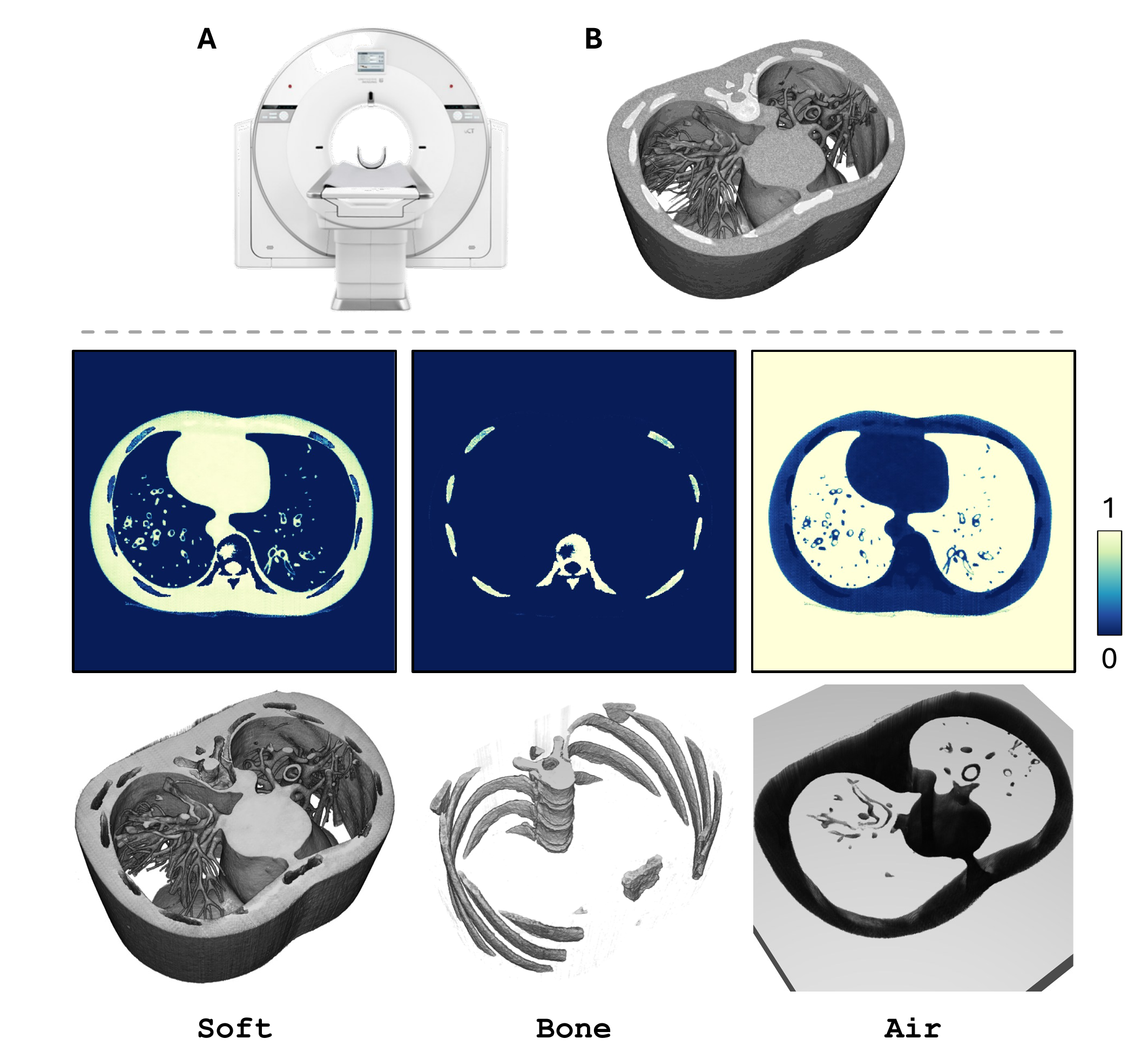}
    \caption{Qualitative results of SEMMD reconstructions using our \roundtworevision{JSolver}-INR on the 3D real-world human body phantom. Row $\#$1 shows the experimental setup, where \textbf{A)} The commercial UIH uCT 768 scanner, and \textbf{B)} The measured CT volume. Rows $\#$2 and $\#$3 illustrate the 2D slice and 3D volume rendering of the SEMMD reconstructions, respectively.}
    \label{fig:simu_c}
\end{figure}
\subsection{Results on Real-world Solution Phantom}
\label{sec:exp_real_solution}
\par Fig.~\ref{fig:fig_real_solution} presents the qualitative results. Due to the presence of noise in the \revision{FBP SECT reconstruction} (Fig.~\ref{fig:real_setup}\textbf{D}), the reconstructions produced by TMA~\cite{xue2020image} exhibit pronounced noise amplification. MSC~\cite{xue2021multi} effectively suppresses this noise through a TV-based smoothness regularization. However, both methods fail to deliver satisfactory SEMMD results. Specifically, for solutions $\#$1 and $\#$4, composed of H$_2$O and CaCl$_2$ in a 0.6$:$0.4 ratio, the decomposition results from TMA~\cite{xue2020image} are dominated by H$_2$O, while MSC~\cite{xue2021multi} further degrades performance due to over-smoothing. The quantitative comparisons are shown in Table~\ref{tab:real_solution}. All three compared methods perform well on the pure-water solutions $\#$2 and $\#$3, owing to their simple composition. However, for the mixed solutions $\#$1 and $\#$4, both TMA and MSC yield results that deviate significantly from the ground truth. In contrast, our \roundtworevision{JSolver}-INR achieves consistently superior SEMMD performance, both in qualitative visualization and quantitative metrics.
\par We also present the results of spectrum estimation in Fig.~\ref{fig:fig_spectrum_pre_solution}. Note that TMA~\cite{xue2020image} and MSC~\cite{xue2021multi} are excluded from this comparison as they do not account for the energy spectrum during the SEMMD process. The reference spectrum is obtained using the SPEKTR toolkit~\cite{punnoose2016technical} based on known X-ray source configurations. As shown, the spectrum estimated by our \roundtworevision{JSolver}-INR closely matches the reference, further validating the effectiveness of our method in real-world spectrum estimation.
\begin{figure}[t]
    \centering
    \includegraphics[width=0.9\linewidth]{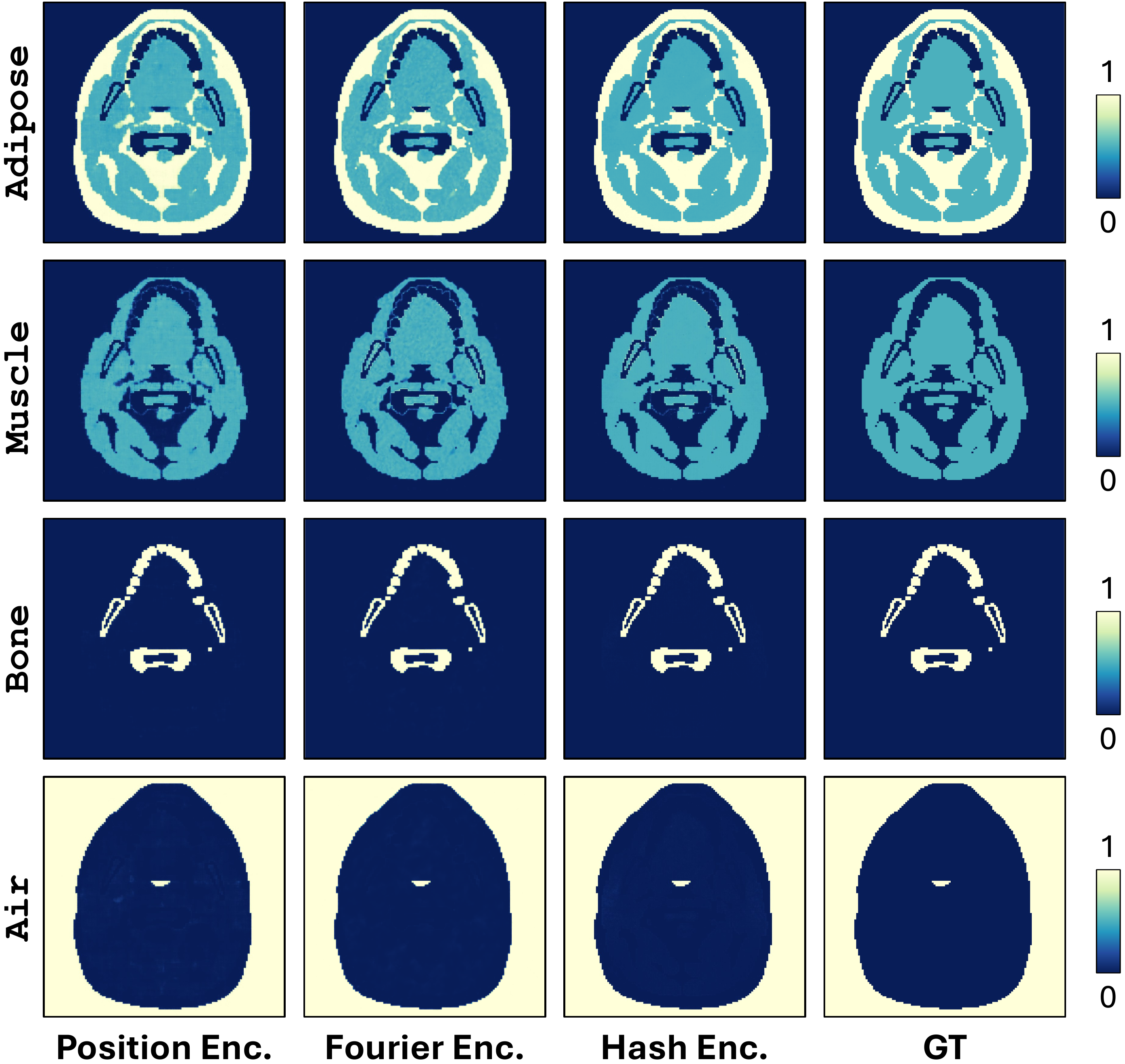}
    \caption{Qualitative comparison of SEMMD results by \roundtworevision{JSolver}-INR model with different network architectures on simulated XCAT phantom \textbf{\texttt{A}}.}
    \label{fig:simu_ab_network}
\end{figure}
\begin{table}[t]
    \centering
    \caption{Quantitative results of \roundtworevision{JSolver}-INR model with different network architectures on two simulated XCAT phantoms. The best performances are highlighted in \textbf{bold}.}
    \tablestyle
    \begin{tabular}{cccc} 
    \toprule
    \multirow{2}{*}{\textbf{Network}} & \textbf{SEMMD} & \textbf{Spectrum} & \textbf{Time}\\ \cmidrule(lr){2-4}
    & \textbf{RMSE} & \textbf{MAE}  & \textbf{Mins} \\\midrule
    Position Enc.~\cite{mildenhall2021nerf} & 0.0320$\pm$0.0071 & \textbf{0.0090$\pm$0.0007} & 7\\ 
    Fourier Enc.~\cite{tancik2020fourier}& 0.0367$\pm$0.0068 & 0.0110$\pm$0.0005&4\\
    Hash Enc.~\cite{hash}& \textbf{0.0206$\pm$0.0066} & 0.0091$\pm$0.0018 & 2\\
    \bottomrule
    \end{tabular}
    \label{tab:ab_newtork}
\end{table}

\begin{figure}[t]
    \centering
    \includegraphics[width=\linewidth]{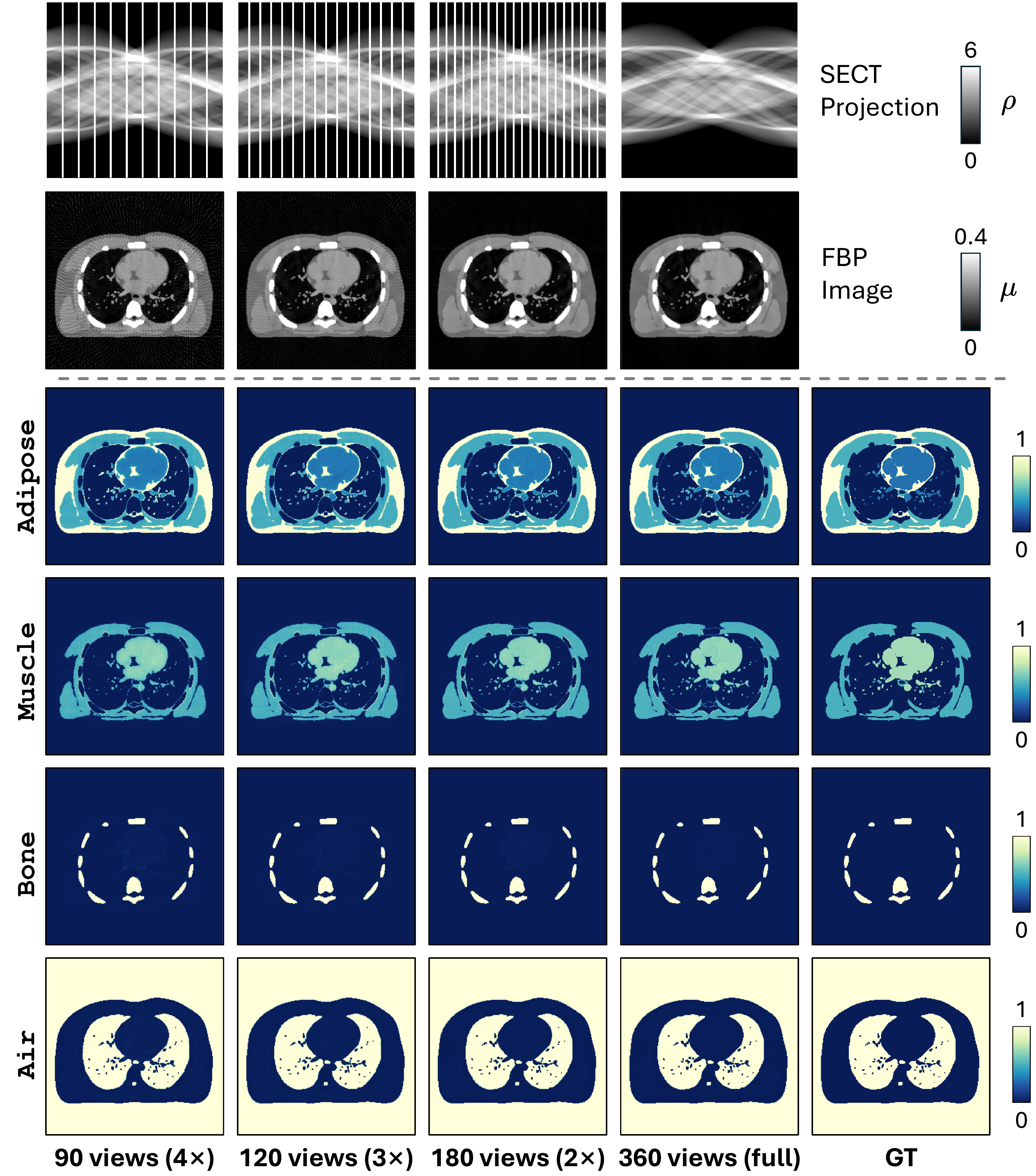}
    \caption{Qualitative comparison of SEMMD results by \roundtworevision{JSolver}-INR model on simulated XCAT phantom \textbf{\texttt{B}} for undersampled CT acquisitions.}
    \label{fig:fig_simu_sub}
\end{figure}
\begin{table}[t]
    \centering
    \caption{Quantitative results of \roundtworevision{JSolver}-INR model on two simulated XCAT phantoms for undersampled CT acquisitions. The best performances are highlighted in \textbf{bold}.}
    \tablestyle
    \begin{tabular}{cccc} 
    \toprule
    \multirow{2}{*}{\textbf{\# Projections}} & \textbf{SEMMD} & \textbf{Spectrum} & \textbf{Time}\\ \cmidrule(lr){2-4}
    & \textbf{RMSE} & \textbf{MAE}  & \textbf{Mins} \\\midrule
    360 views (full) & \textbf{0.0206$\pm$0.0066} & \textbf{0.0091$\pm$0.0018} & 2\\
    180 views (2$\times$)& 0.0293$\pm$0.0061 & 0.0116$\pm$0.0019 & {1}\\
    120 views (3$\times$)& 0.0360$\pm$0.0066 & 0.0242$\pm$0.0038 & {0.7}\\
    90 views (4$\times$)& 0.0413$\pm$0.0079 & 0.0249$\pm$0.0019 & \textbf{0.5}\\
    \bottomrule
    \end{tabular}
    \label{tab:sub}
\end{table}

\subsection{Results on Real-world Human Body Phantom}
\label{sec:exp_real_human}
To evaluate our method in clinical settings, we used a commercial UIH uCT 768 scanner under clinical helical CBCT protocols to scan a human body phantom. The experimental setup is shown in Fig.\ref{fig:simu_c} (\textbf{A}\&\textbf{B}). \revision{TMA~\cite{xue2020image} and MSC~\cite{xue2021multi} are established optimization-based SEMMD methods, but their original algorithms perform 2D image-domain voxel-wise optimization and do not support the full helical projection geometry. Extending them to a 512$\times$512$\times$175 volume also entails substantial computational and memory costs. Recent supervised SEMMD methods~\cite{zhao2019deep,zhao2020deep,cong2020virtual,yuasa2023pseudo} mainly use 2D networks. Their direct extension to 3D convolution is computationally demanding and requires paired multi-energy CT measurements or material labels, which are unavailable for this human phantom. Therefore, the 3D experiment is presented as a feasibility validation, while the quantitative baseline comparisons are conducted on the controlled 2D datasets.} We assume the human body phantom is composed of three base materials: soft tissue, bone, and air. Here, we present qualitative SEMMD results in Fig.~\ref{fig:simu_c}. Quantitative metrics are not provided because the GT decomposition maps could not be scanned. Additionally, the estimated spectrum is not shown due to commercial privacy concerns.
\par As shown in Fig.~\ref{fig:simu_c}, our \roundtworevision{JSolver}-INR model produces high-quality decomposition maps, capturing both global structure and local anatomical details. This study, based on clinical CT scanners, demonstrates the reliability of our method.
\par \revision{For each 3D optimization step, the dominant computational complexity is approximately $O(|\mathbf{R}|N_sML)$, where $|\mathbf{R}|$ is the number of sampled rays, $N_s$ is the number of points sampled along each ray, $M$ is the number of basis materials, and $L$ is the number of energy bins. The full 3D reconstruction uses 3,000 epochs, with 200 rays per step and 1,533 points per ray, and requires approximately 40 minutes on an NVIDIA RTX 4070 Ti Super GPU. Under the same tensor dimensions and batch configuration, the observed peak GPU memory and peak process RAM are 15.55 GB and 9.13 GB, respectively. Peak memory is measured using one optimization epoch followed by one full-volume inference, which covers the memory-intensive training and reconstruction operations without repeating all epochs.}
\subsection{Impact of Network Architecture on \roundtworevision{JSolver}}
\label{sec:exp_ab_network}
\par Our network $\mathcal{F}_\Phi$ is implemented using hash encoding~\cite{hash} and a two-layer MLP. On the two simulated XCAT phantoms, we investigate the impact of different network architectures on model performance, including SEMMD accuracy, spectrum estimation, and runtime. Specifically, we compare three types of architectures: hash encoding with a two-layer MLP, position encoding\cite{mildenhall2021nerf} with a six-layer MLP, and Fourier encoding~\cite{tancik2020fourier} with a six-layer MLP. All other model configurations are kept the same to ensure a fair comparison.
\par Table~\ref{tab:ab_newtork} presents the quantitative results. All three architectures achieve comparable and strong performance in both SEMMD accuracy and spectrum estimation. However, the hash encoding~\cite{hash} offers the highest computational efficiency, requiring only 2 minutes. We also demonstrate the SEMMD reconstructions in Fig.~\ref{fig:simu_ab_network}, where all three architectures produce high-quality decomposed maps. 
\par This ablation study on network architecture suggests that the proposed \roundtworevision{JSolver} model is not highly sensitive to the choice of network architecture for SEMMD and spectrum estimation. Nevertheless, exploring more advanced architectures may further accelerate model optimization.
\subsection{Performance of \roundtworevision{JSolver} with INR for Undersampled CT}
\label{sec:exp_undersampled}
\par Benefitting from the continuous priors provided by neural networks~\cite{rahaman2019spectral}, INR enables high-quality CT reconstructions from undersampled acquisitions. Our \roundtworevision{JSolver}-INR model, empowered by INR, is therefore expected to perform robustly under undersampled SECT projections. To evaluate this, we vary the number of CT projection views. Specifically, we use the two simulated XCAT phantoms of size 256$\times$256, with 360 views considered as the fully sampled baseline. We then set three undersampling levels: \{180, 120, 90\} views, corresponding to undersampling rates of {2$\times$, 3$\times$, 4$\times$}, respectively. The qualitative results are presented in Fig.~\ref{fig:fig_simu_sub}. The top two rows show that as the number of projection views decreases, SECT images reconstructed using FBP~\cite{fbp} exhibit increasingly severe streak artifacts. Nevertheless, our \roundtworevision{JSolver}-INR model consistently yields high-quality SEMMD reconstructions across all scenarios. Quantitative results are summarized in Table~\ref{tab:sub}. As expected, the performance on both SEMMD and spectrum estimation degrades with fewer projection views. However, we emphasize that despite this degradation, the model maintains a consistently high level of accuracy. This study demonstrates that INR significantly enhances the robustness of our \roundtworevision{JSolver} model, enabling reliable reconstructions even under challenging undersampled CT acquisition settings.

\section{Conclusion}
\par In this work, we proposed \roundtworevision{JSolver}, a novel one-step SEMMD optimization framework. By explicitly incorporating physics-informed spectral priors into the SEMMD process, \roundtworevision{JSolver} accurately simulates energy-dependent CT acquisitions. As a result, it enables joint reconstruction of multi-material decomposition and estimation of the X-ray energy spectrum directly from raw SECT projections, without requiring any external data. Furthermore, we introduce INR as a powerful unsupervised deep learning solver to model the decomposition maps, significantly enhancing SEMMD reconstruction quality. \revision{To the best of our knowledge, \roundtworevision{JSolver} is the first unsupervised INR-based approach designed for both SEMMD and spectrum estimation.} Extensive evaluations demonstrate its superiority over existing cutting-edge techniques in terms of both reconstruction accuracy and computational efficiency. \revision{The current experiments do not fully validate high-Z K-edge contrast-agent decomposition, such as iodine or gadolinium; although the forward model can incorporate their energy-dependent attenuation curves, such materials may create stronger spectrum-material coupling and will be investigated in future work.}

\bibliography{ref}
\bibliographystyle{ieeetr}
\end{document}